\documentclass[10pt,twocolumn,letterpaper]{article}

\usepackage[pagenumbers]{cvpr} 

\definecolor{cvprblue}{rgb}{0.21,0.49,0.74}
\usepackage[pagebackref,breaklinks,colorlinks,allcolors=cvprblue]{hyperref}
\usepackage{multirow, multicol}
\usepackage{diagbox}
\usepackage{pifont}
\usepackage{lipsum}
\usepackage{tikz} 
\newcommand{\cmark}{\ding{51}}%
\newcommand{\xmark}{\ding{55}}%
\usepackage{colortbl, xcolor}
\usepackage{bbding}
\usepackage{algorithm}
\usepackage{algpseudocode}
\usepackage{amsmath}
\definecolor{Yellow}{rgb}{1,1,0.80}
\newcommand{\redxmark}{{\color{red} \ding{55}}} 
\newcommand{\greencheck}{{\color{ForestGreen} \ding{51}}} 
\newcommand\blfootnote[1]{%
  \begingroup
  \renewcommand\thefootnote{}\footnote{#1}%
  \addtocounter{footnote}{-1}%
  \endgroup
}
\usepackage{amsmath}
\newcommand{\doubleunderline}[1]{\underline{\underline{#1}}}
\title{Efficient Personalization of Quantized Diffusion Model without Backpropagation}

\author{Hoigi Seo$^{1*}$ \qquad Wongi Jeong$^{1*}$ \qquad Kyungryeol Lee$^{1}$ \qquad Se Young Chun$^{1,2\dag}$ \\
$^1$Dept. of Electrical and Computer Engineering, $^2$INMC \&  IPAI \\
Seoul National University, Republic of Korea \\
{\tt\small \{seohoiki3215, wg7139, kr.lee, sychun\}@snu.ac.kr}
}

\begin{document}
\maketitle
\blfootnote{* Authors contributed equally. $\dag$ Corresponding author.}
\begin{abstract}

Diffusion models have shown remarkable performance in image synthesis, but they demand extensive computational and memory resources for training, fine-tuning and inference. Although advanced quantization techniques have successfully minimized memory usage for inference, training and fine-tuning these quantized models still require large memory possibly due to dequantization for accurate computation of gradients and/or backpropagation for gradient-based algorithms. However, memory-efficient fine-tuning is particularly desirable for applications such as personalization that often must be run on edge devices like mobile phones with private data. In this work, we address this challenge by quantizing a diffusion model with personalization via Textual Inversion and by leveraging a zeroth-order optimization on personalization tokens without dequantization so that it does not require gradient and activation storage for backpropagation that consumes considerable memory. Since a gradient estimation using zeroth-order optimization is quite noisy for a single or a few images in personalization, we propose to denoise the estimated gradient by projecting it onto a subspace that is constructed with the past history of the tokens, dubbed Subspace Gradient. In addition, we investigated the influence of text embedding in image generation, leading to our proposed time steps sampling, dubbed Partial Uniform Timestep Sampling for sampling with effective diffusion timesteps. Our method achieves comparable performance to prior methods in image and text alignment scores for personalizing Stable Diffusion with only forward passes while reducing training memory demand up to $8.2\times$. 
Project page: \url{https://ignoww.github.io/ZOODiP_project/}

\end{abstract}    
\section{Introduction}
\label{sec:intro}
Recent advances in diffusion models~\cite{ho2020denoising, sohl2015deep, song2019generative, song2020score, watson2022learning} have revolutionized generative AI, offering a powerful framework for high-quality, diverse data generation. Diffusion models have shown impressive capabilities across various applications, including image synthesis~\cite{ho2020denoising, song2020denoising, peebles2023scalable, liu2022flow, song2019generative, sohl2015deep, watson2022learning}, text-guided image synthesis~\cite{rombach2022high, esser2024scaling, chen2024pixart, podell2023sdxl, ramesh2022hierarchical}, video generation~\cite{blattmann2023stable, liu2024sora, ho2022video, kim2024versatile, ho2022imagen}, and 3D content generation~\cite{seo2023ditto, poole2022dreamfusion, lin2023magic3d, liu2023zero, wang2023score}, outperforming traditional generative approaches. However, modern diffusion models face significant memory and computational costs during training, fine-tuning, and inference due to their size.

These overheads pose major challenges in personalization tasks, which require fine-tuning diffusion models with only a few user-provided images. Typically, personalization is achieved by either fine-tuning the denoising network~\cite{ruiz2023dreambooth, kumari2023multi, tewel2023key, xiang2023closer, gu2024mix, ruiz2024hyperdreambooth} or introducing new text tokens~\cite{gal2022image, kumari2023multi, park2024textboost, voynov2023p+, alaluf2023neural, zhang2023prospect} to represent desired subjects. However, both approaches require gradient-based optimization, leading to significant memory and computational overhead, which is especially problematic in memory-constrained on-device training for sensitive data.
To tackle the overheads in personalization, efficient tuning techniques have emerged. Existing methods reduce trainable parameters~\cite{kumari2023multi,tewel2023key, xiang2023closer, gu2024mix, ruiz2024hyperdreambooth}, leverage quantized models~\cite{ryu2024memory, kim2024memory}, or employ gradient-free optimization with evolution strategy~\cite{fei2023gradient}. However, these methods have limitations: 1) they mostly rely on backpropagation, unsuitable for most mobile processors, which are designed to accelerate inference~\cite{qualcomm2024unlock}; 2) they still require significant memory for storing activations and gradients; 3) evolutionary algorithms can be unstable and inefficient in small-batch.

\begin{figure*}[h]
  \centering
  \begin{subfigure}{0.59\linewidth}
    \includegraphics[width=0.9\linewidth]{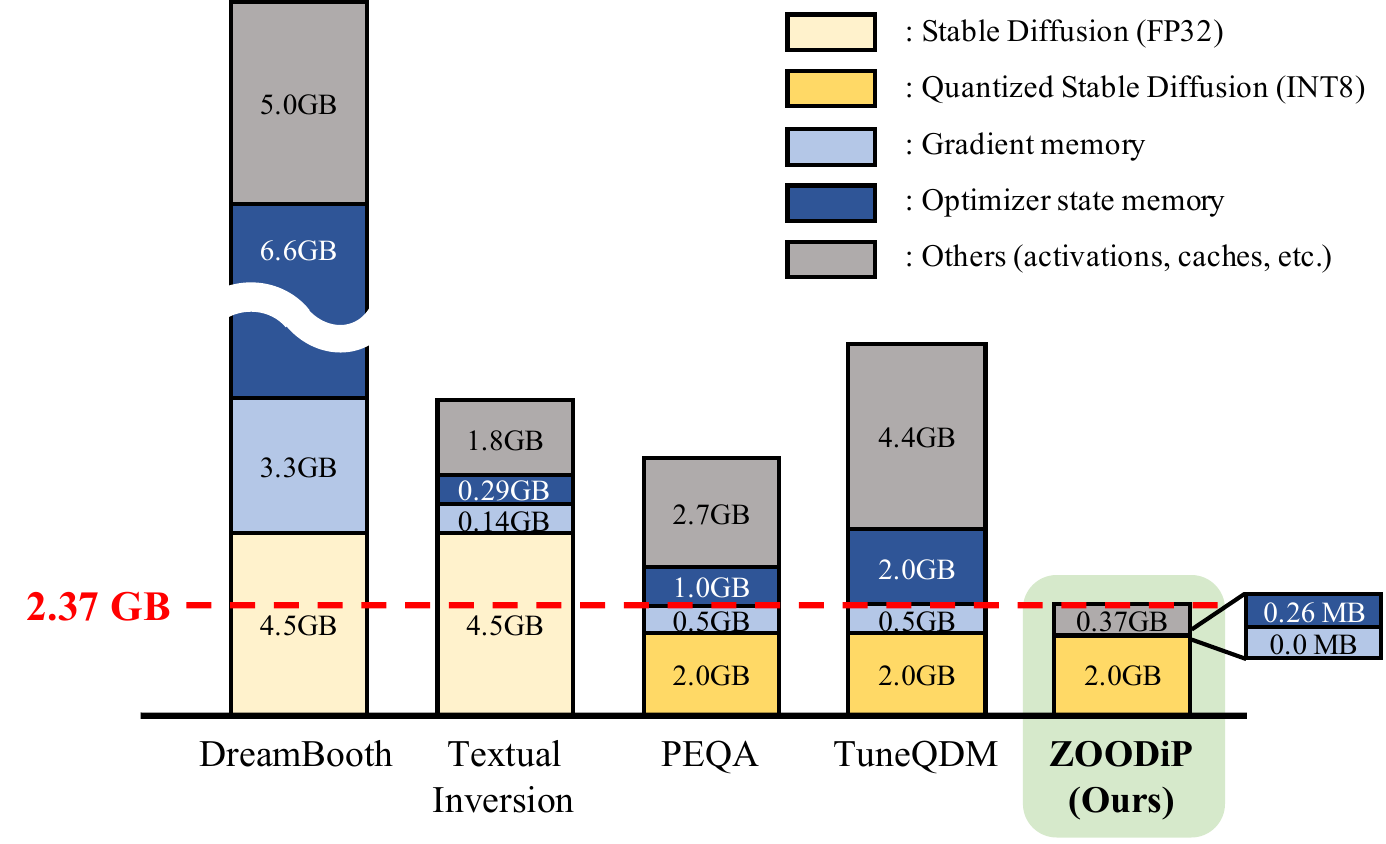}
    \caption{GPU memory breakdown across various personalization methods.}
    \label{fig:short-a}
  \end{subfigure}
  \begin{subfigure}{0.40\linewidth}
    \includegraphics[width=\linewidth]{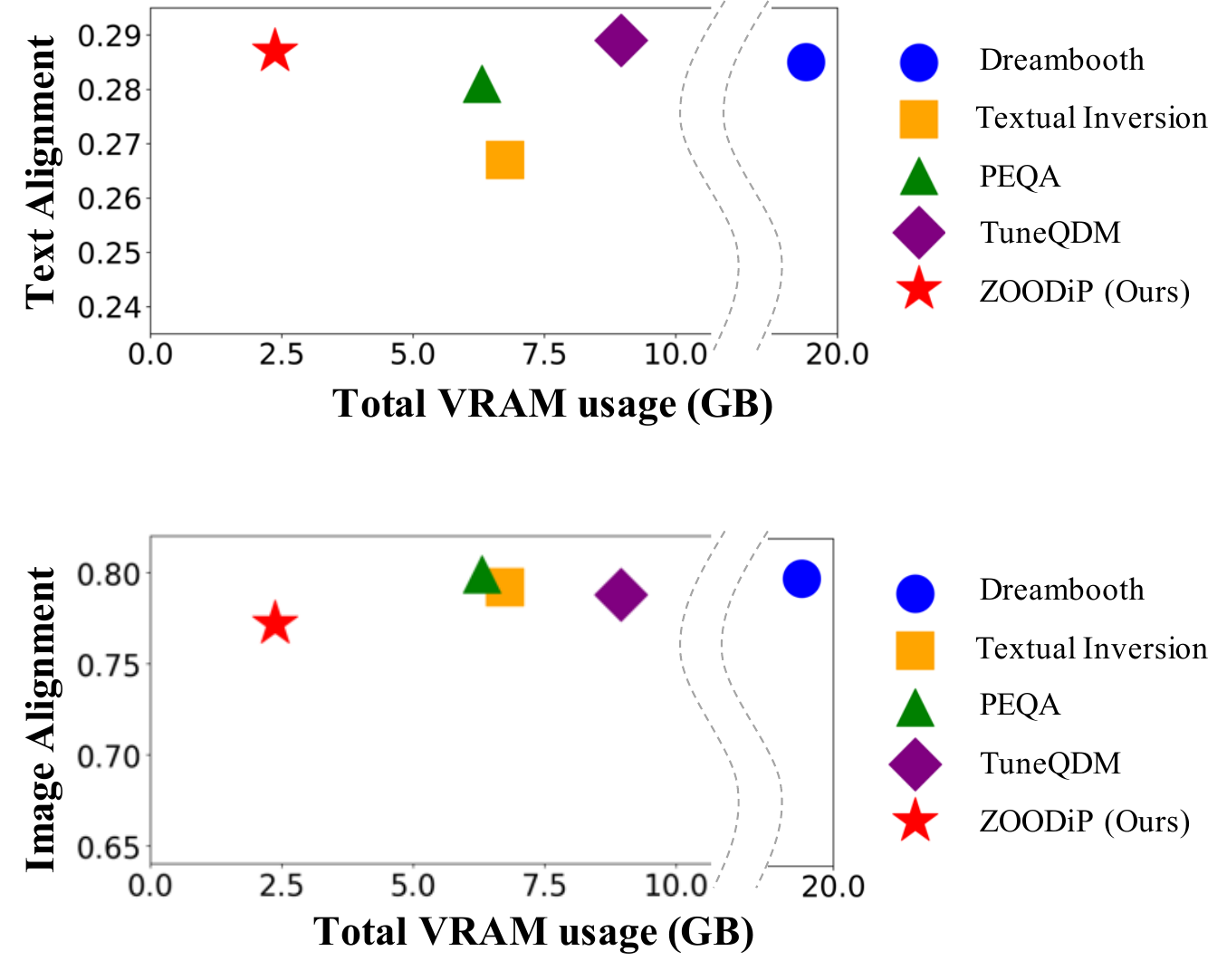}
    \caption{VRAM usage versus image and text alignment scores.}
    \label{fig:short-b}
  \end{subfigure}
  \caption{Analysis of memory consumption and performance of Stable Diffusion personalization methods. \textbf{(Left)} GPU memory breakdown for each method on a Stable Diffusion personalization with a batch size of 1. ZOODiP (Ours) shows significantly higher memory efficiency compared to other methods. \textbf{(Right)} Comparison of memory usage versus performance across methods. Performance is measured with text (CLIP-T) and image (CLIP-I) alignment scores. ZOODiP achieves comparable performance to other methods while using significantly less memory (up to $8.2\times$ less than DreamBooth). Memory usage was profiled using the PyTorch profiler and \texttt{nvidia-smi} command.}
  \label{fig:fig1_main_result}
\end{figure*}

Here we propose Zeroth-Order Optimization for Diffusion model Personalization (ZOODiP) that can personalize Stable Diffusion with 2.37GB VRAM consumption using only forward passes without compromising the image quality.
Our method relies on three key observations: First, Zeroth-Order (ZO) optimization effectively handles non-differentiable objectives~\cite{malladi2023fine} (\textit{e.g.}, accuracy) during training. Second, tokens optimized via Textual Inversion~\cite{gal2022image} undergo significant changes in a low-dimensional subspace. Principal Component Analysis (PCA) on the token embeddings reveals that the initial and personalized tokens primarily update within this subspace. Third, based on prior works~\cite{balaji2022ediff,go2024addressing,lee2024multi, liu2023oms, patashnik2023localizing, choi2022perception, chefer2023attend, zhang2023prospect, daras2022multiresolution}, which argue that timesteps have distinct roles in diffusion models, we identified an effective timestep section for personalization.

Following the first observation, we trained token embeddings using ZO optimization in a quantized, non-differentiable model~\cite{malladi2023fine, yin2019understanding, bengio2013estimating}. This method reduces memory usage for activations, weights, and computational costs on edge devices. Inspired by the second finding, we introduced Subspace Gradient (SG) to accelerate training by mitigating noisy gradients. SG projects out dimensions with noisy gradients based on parameter trajectory, improving performance. Based on the third observation, we applied Partial Uniform Timestep Sampling (PUTS) within targeted timestep sections, skipping less influential timesteps to maximize impact within fixed training iterations.

We comprehensively assessed ZOODiP using both qualitative and quantitative measures on DreamBooth~\cite{ruiz2023dreambooth} dataset. We measured performance across two key metrics: 1) text-image alignment and 2) reference-generated image alignment. Through GPU memory profiling during training, we demonstrate that ZOODiP utilizes up to $\mathbf{8.2\times}$ \textbf{less memory} than existing methods, achieving similar image quality. This substantial reduction in memory footprint underscores ZOODiP's efficacy in enabling diffusion model personalization on memory-constrained devices.

Our contributions can be summarized as follows:
\begin{itemize}
    \item We empirically identify that Textual Inversion tokens primarily optimize within a low-dimensional subspace, and that focusing on partial timesteps accelerates training.
    \item We introduce ZOODiP, a novel method that combines ZO optimization with a quantized model, subspace gradient projection, and customized timestep sampling for efficient diffusion model personalization.
    \item We demonstrate that ZOODiP achieves competitive performance with significantly lower memory requirements on the DreamBooth dataset (see Fig.~\ref{fig:fig1_main_result}).
\end{itemize}


\section{Related Works}
\label{sec:related_works}

\begin{figure*}[!h]
  \centering
  \begin{subfigure}{0.6\linewidth}
    \includegraphics[width=1.0\linewidth]{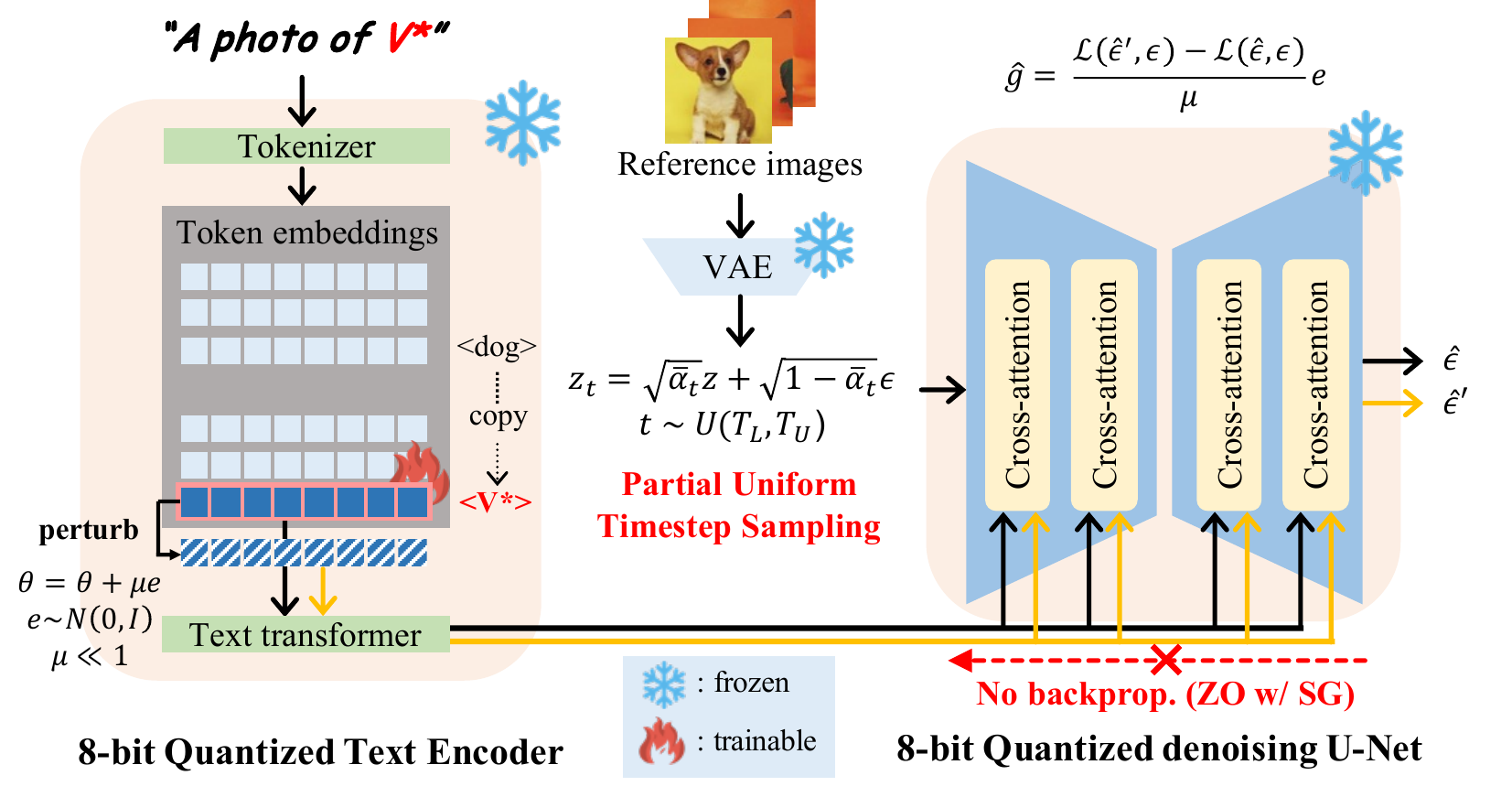}
    \caption{Overall illustration of ZOODiP training framework.}
    \label{fig:method-a}
  \end{subfigure}
  \begin{subfigure}{0.36\linewidth}
    \includegraphics[width=1.0\linewidth]{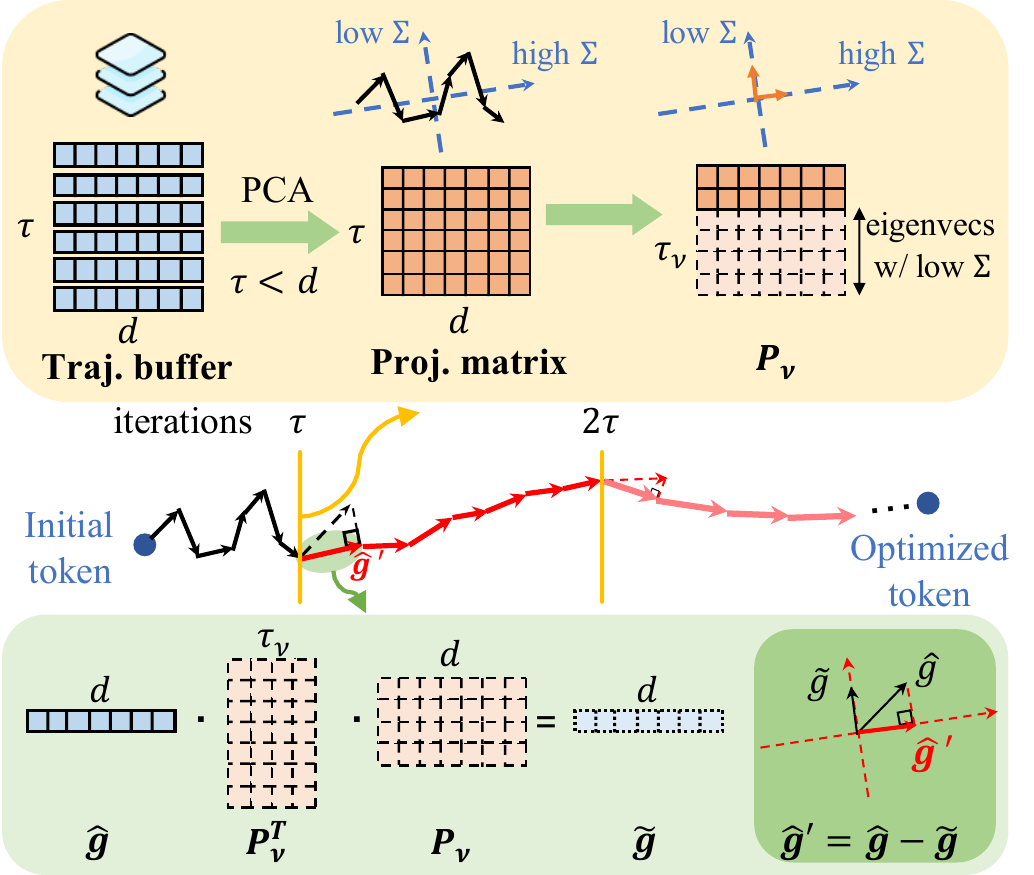}
    \caption{Illustration of Subspace Gradient (SG) updates.}
    \label{fig:method-b}
  \end{subfigure}
  \caption{\textbf{(a)} Illustration of overall ZOODiP framework. A target token is initialized and added to the prompt. Reference images are encoded, and Partial Uniform Timestep Sampling (PUTS)-sampled timestep noise is predicted. The loss is calculated with the original and perturbed token to estimate the gradient. \textbf{(b)} Illustration of Subspace Gradient (SG). Updated tokens from the previous $\tau$ iterations are stored. PCA identifies low-variance eigenvectors to project out noisy dimensions from the estimated gradient for the next $\tau$ iterations.}
  \label{fig:method}
\end{figure*}

\paragraph{Diffusion model personalization.}
Diffusion model personalization aims to adapt a pre-trained model to generate images of new, user-defined concepts using a small set of provided images. Textual Inversion~\cite{gal2022image} tackles personalization by optimizing a single token embedding that represents the target concept. DreamBooth~\cite{ruiz2023dreambooth} tunes the denoising U-Net, while Custom Diffusion~\cite{kumari2023multi} personalizes multiple concepts by adopting new tokens and adjusting the key and value matrices in cross-attention. $\mathcal{P}+$~\cite{voynov2023p+} assigns unique text tokens to each U-Net stage, and TextBoost~\cite{park2024textboost} introduces an augmentation token and employs SNR-based sampling for single-image personalization. However, these methods often require significant computational resources and memory, limiting their applicability on resource-constrained devices. Inspired by the efficiency of single-token optimization in Textual Inversion, ZOODiP introduces a new token to represent the target concept and optimizes it efficiently.

\paragraph{Efficient fine-tuning.}
Fine-tuning large models requires significant memory, motivating the memory-efficient training methods. Many approaches aim to reduce the number of trainable parameters. For example, LoRA~\cite{hu2021lora} applies low-rank matrices to linear layers, effectively decreasing the number of parameters to update. QLoRA~\cite{dettmers2024qlora} further reduces memory usage by applying LoRA to quantized models. Another direction explores fine-tuning quantized models by adjusting quantization parameters. PEQA~\cite{kim2024memory} only tunes the quantization scales to reduce the optimizer state memory. TuneQDM~\cite{ryu2024memory} personalizes quantized diffusion models by tuning the scales specific to each diffusion timestep set. Beyond parameter reduction and quantization, Gradient-Free Textual Inversion~\cite{fei2023gradient} employs an evolution strategy to optimize tokens, bypassing backpropagation.

However, these techniques can be limited by memory-intensive backpropagation or training instability. ZOODiP overcomes these limitations by utilizing zeroth-order optimization on a quantized model, eliminating backpropagation and its associated memory overhead.

\paragraph{Zeroth-order optimization.}
As model sizes increased, memory consumption during backpropagation became a major challenge for fine-tuning. Zeroth-order (ZO) optimization~\cite{spall1992multivariate, liu2020primer, nesterov2017random, ghadimi2013stochastic, duchi2015optimal, bollapragada2018adaptive, cai2022zeroth}, which estimates gradients using only forward passes with random perturbations, has emerged as a promising solution. MeZO~\cite{malladi2023fine} demonstrated memory efficient tuning of large language models with forward-pass only, storing random seeds instead of perturbation vectors. Additionally, MeZO showed that ZO's convergence rate depends on the model's effective rank and that non-differentiable objectives can be optimized with ZO. Similarly, DeepZero~\cite{chen2023deepzero} trained models from scratch using parallelized ZO, achieving performance comparable to first-order methods. Inspired by these advancements, we propose a personalization framework that employs ZO optimization on a quantized model, significantly reducing memory overhead for both activations and weights.
\section{Method}
\label{sec:method}
In this section, we propose ZOODiP, a memory-efficient approach to personalize diffusion models. It leverages zeroth-order (ZO) optimization with a quantized model, eliminating the need for backpropagation, and thereby significantly reducing memory usage. Based on the observation that trained tokens in Textual Inversion primarily change within a low-dimensional subspace, we introduce Subspace Gradient (SG) to optimize tokens within this reduced space. Furthermore, we propose Partial Uniform Timestep Sampling (PUTS) for efficient timestep selection, capitalizing on the observation that text embeddings predominantly influence image generation at specific diffusion timesteps. By integrating ZO optimization with quantization, SG, and PUTS, ZOODiP enables efficient personalization on resource-constrained devices. The overall framework is illustrated in Fig.~\ref{fig:method} and Algorithm.~\ref{alg:ZOODiP}.

\subsection{Formulation}
ZOODiP builds upon Textual Inversion, aiming to learn a pseudo-token $v^*$ that represents a desired concept from a set of reference images ${x_1,...,x_n}\in\mathcal{X}$ containing that concept. This pseudo-token is incorporated into the model's text embedding. We denote the condition for denoising network $y^*$ as the text prompt that includes $v^*$. Following the previous studies~\cite{rombach2022high, gal2022image}, we formulate the objective as follows:
\begin{equation} \label{eq:ldm_loss}
    \begin{gathered}
        \mathcal{L}_{\text{LDM}} = \mathbb{E}_{z\sim\mathcal{E}(x),y^{*},\epsilon\sim\mathcal{N}(0,I),t}\big[||\epsilon - \epsilon_{\phi}(z_t, t, c(y^*))||_2^2\big], \\
        z_t = \sqrt{\bar{\alpha_t}}z + \sqrt{1-\bar{\alpha_t}}\epsilon
    \end{gathered}
\end{equation}
where $x$ is the input reference image, $\mathcal{E}$ is a Variational Autoencoder (VAE) encoder, $z$ is the latent encoded with the encoder, $\epsilon$ is the noise from the forward process, $\epsilon_{\phi}$ is a denoising network, $t$ is the diffusion timestep, $c$ is a condition encoder which is text encoder in ZOODiP, and $\bar{\alpha}_t$ is cumulative product of $\alpha$ from $0$ to $t$ as defined in DDPM~\cite{ho2020denoising}.

\begin{algorithm}[t]
\caption{Fine-tuning algorithm of ZOODiP}
\label{alg:ZOODiP}
\begin{algorithmic}[1]
\Require token embedding $\theta$, reference image set $\mathcal{X}$, text input $y^{*}$, text encoder $c$, VAE encoder $\mathcal{E}$, denoising network $\epsilon_{\phi}$, buffer size $\tau$, threshold $\nu$, number of estimation $n$, perturbation size $\mu$, total iteration $L$, trajectory buffer $B$, learning rate $\eta$
\State $P_{\nu}\gets\mathbf{0}$
\For{$l = 1, \dots, L$}
    \State Sample image $x\in\mathcal{X}$
    \State $t\sim U(T_L,T_U)$ \Comment{PUTS}
    \State $\hat{g}_{\theta}\gets$ RGE($x,y^{*},n,t,\mathcal{E},c,\epsilon_{\phi},\mu$)\Comment{Eq.~\ref{eq:ldm_loss} and Eq.~\ref{eq:RGE}}
    \State $\hat{g}'_{\theta}\gets \hat{g}_{\theta}(I - P_{\nu}^\top P_{\nu})$  \Comment{SG update}
    \State $\theta\gets$ optimizer\,($\theta,\hat{g}'_{\theta},\eta$)
    \State $B_{\text{mod}(l,\tau)}\gets\theta$ \Comment{Fill the buffer with updated token}
    \If{$\text{mod}(l,\tau) = 0$} \Comment{Update $P_{\nu}$ for SG}
        \State $P_{\nu}\gets$update$P_\nu$($B,\nu$) \Comment{Algorithm.~\ref{alg:subspace}}
    \EndIf
\EndFor
\end{algorithmic}
\end{algorithm}

\begin{algorithm}[t]
\caption{Subspace generation}
\label{alg:subspace}
\begin{algorithmic}[1]
\Function{update$P_\nu$}{$B,\nu$}
    \State $\bar{B}\gets$Normalize($B$) \Comment{Eq.~\ref{eq:buffer_normalize}}
    \State $\lambda$, $V \gets \text{PCA}(\bar{B})$
    \State $i^* \gets \arg\min_{i}\left\{ i \in \{1: \tau\} \mid \frac{\sum_{j=1}^{i} \lambda_j}{\sum_{k=1}^{\tau} \lambda_k} > 1 - \nu \right\}$
    \State $P_{\nu}\gets V^\top_{i^{*}:\tau}$
    \State \Return $P_{\nu}$
\EndFunction
\end{algorithmic}
\end{algorithm}

\subsection{ZO Optimization with Quantized Model}
To optimize the $v^*$ with memory efficiency, we quantize the weights in the Linear and Convolution layer of all components in the diffusion model: U-Net, VAE, and the text encoder. The quantization is formulated as follows~\cite{krishnamoorthi2018quantizing}:
\begin{equation}
\label{eq:quant}
\widetilde{W} = s \cdot \left(\operatorname{clamp}\left(\left\lfloor\frac{W}{s}\right\rceil + o, Q_N, Q_P\right) - o\right)
\end{equation}
where $W$ is a weight to be quantized, $\lfloor \cdot \rceil$, $s$, $o$ represent the rounding function, quantization scale, and zero-point, respectively. The minimum and maximum quantization bounds, $Q_N=-2^{N-1}$ and $Q_P=2^{N-1}-1$, are configured to ensure symmetric $N$-bit quantization. After the quantization of the network, $\mathcal{L}_{\text{LDM}}$ in Eq.~\ref{eq:ldm_loss} becomes non-differentiable due to the existence of the rounding function $\lfloor \cdot \rceil$, which makes training a quantized model difficult~\cite{bengio2013estimating}.

To estimate the gradient on a quantized model, we leverage Random Gradient Estimation (RGE)~\cite{spall1992multivariate, nesterov2017random, chen2023deepzero, cai2022zeroth}. RGE perturbs $\theta\in\mathbb{R}^d$, the token embedding for $v^*$, with random variables and estimates the gradient by calculating the non-differentiable objective $\mathcal{L}_{\text{LDM}}(\theta)$ as follows:
\begin{equation}\label{eq:RGE}
\hat{g}_{\theta} = \frac{1}{n}\sum_{i=1}^{n}\Bigg[ \frac{\mathcal{L}_{\text{LDM}}(\theta + \mu e_i) - \mathcal{L}_{\text{LDM}}(\theta)}{\mu}e_i \Bigg] \end{equation}
where $\hat{g}_{\theta}$ denotes an estimation of First-Order (FO) gradient $\nabla_{\theta}\mathcal{L}(\theta)$ with respect to $\theta$, $\mu$ is a perturbation size, $n$ is the number of random directions to estimate the gradient and $e_i$ a random vector sampled from $\mathcal{N}(0, I)$. With the ZO approach, we were able to estimate the gradient without backpropagation while effectively reducing memory usage.

\begin{figure}[t]
  \centering
  \includegraphics[width=\linewidth]{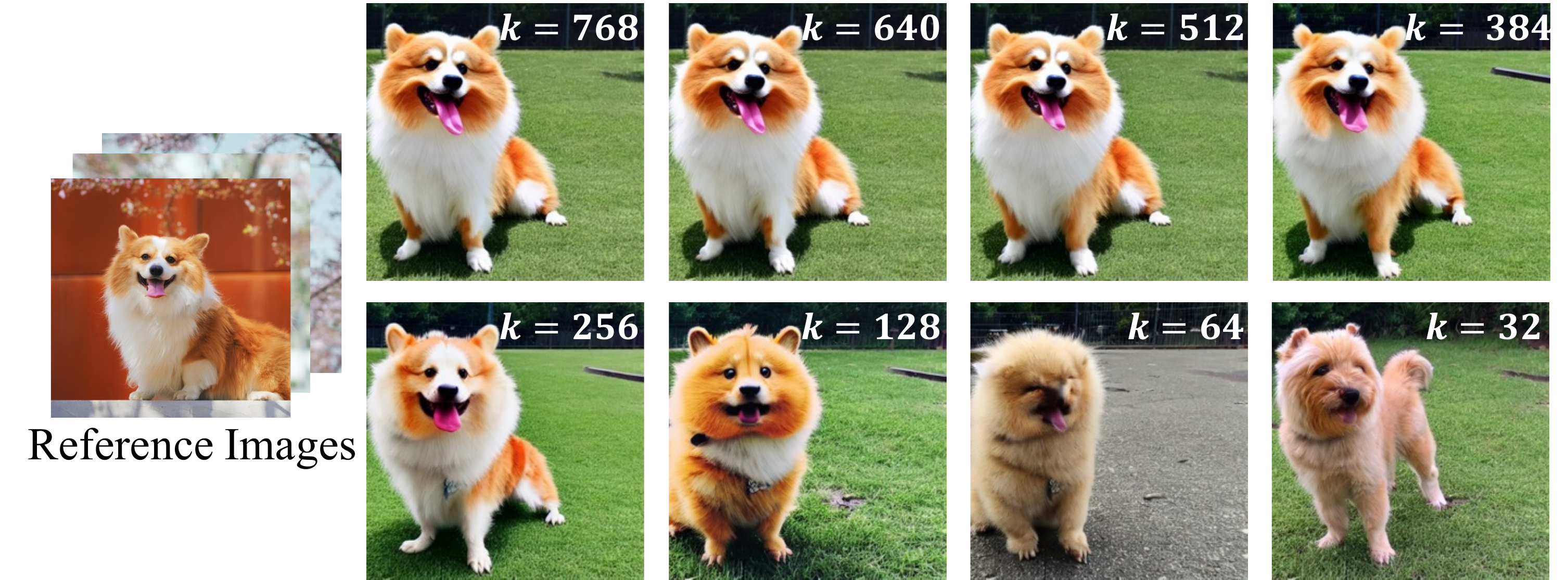}
  \caption{Sparse effective dimension in the token trained with Textual Inversion. Notably, the concept was preserved even when retaining only one-third of the optimized dimensions ($k=256$).}
  \label{fig:subspace}
\end{figure}

\subsection{Subspace Gradient}
\label{subsec:SG}
We investigated the optimization of the pseudo-token by Textual Inversion 
~\cite{gal2022image}(TI) using Principal Component Analysis (PCA) of the token embeddings. Both the initial and optimized tokens were projected onto the principal components. The components with the top-$k$ largest changes in the optimized token were retained, while the rest were replaced with components from the initial token before back-projection. Fig.~\ref{fig:subspace} shows that the personalized concept is preserved after back-projection, indicating that key changes from TI lie within this subspace.

ZO optimization suffers from slow training due to noisy gradient estimates. Inspired by the fact that token optimization mainly occurs in a lower-dimensional subspace, we propose Subspace Gradient (SG) to accelerate training. SG leverages token trajectories to eliminate noisy dimensions. Specifically, we store updated tokens for $\tau$ iterations in a trajectory buffer $B \in \mathbb{R}^{\tau \times d}$, where $d > \tau$. Once $B$ is full, we normalize it as follows:
\begin{equation}\label{eq:buffer_normalize}
    \bar{B}=(B-\mu_B) / \sigma_B
\end{equation}
where $\mu_B$ and $\sigma_B$ is the mean and standard deviation of $B$ along each feature dimension, respectively. Then we perform PCA on $\bar{B}$ with singular value decomposition to get eigenvalues and eigenvectors of the covariance matrix:
\begin{equation}\label{eq:svd}
    \bar{B}=U\Sigma V^\top,\quad\lambda_i=\Sigma_{ii}^2 
\end{equation}
The square of $i$-th diagonal elements of $\Sigma$, denoted as $\lambda_i$, represent the variance explained by the corresponding eigenvectors. We calculate the ratio of the cumulative sum of $\lambda_i$ to the total sum:
\begin{equation}\label{eq:nu_threshold}
    i^* = \arg\min_{i}\left\{i\in\{1,\ldots,\tau\}\Bigg| \dfrac{\sum_{j=1}^{i}\lambda_j}{\sum_{k=1}^{\tau}\lambda_k} > 1-\nu\right\}
\end{equation}
where $\nu$ is a hyperparameter that controls the amount of variance retained. This determines $i^*$, the smallest index $i^*$ for which the cumulative ratio exceeds $1-\nu$. We then construct a projection matrix $P_{\nu}$ using the eigenvectors corresponding to the remaining variance:
\begin{equation}\label{eq:P_nu}
    P_{\nu}:=V^\top_{(i^*+1):\tau}
\end{equation}
This matrix $P_{\nu}$ represents the dimensions to be removed. We project the estimated row vector gradient $\hat{g}$ onto the subspace orthogonal to $P_{\nu}$ and subtract from $\hat{g}$ to get $\hat{g}'$:
\begin{equation}\label{eq:subtract}
    \hat{g}'=\hat{g}(I - P_{\nu}^\top P_{\nu})
\end{equation}
This subtraction effectively eliminates noisy components from the $\hat{g}'$. After updating $P_{\nu}$, the buffer $B$ is cleared. Over the next $\tau$ iterations, the estimated gradient is projected out through $P_{\nu}$, the token trajectory is simultaneously accumulated in $B$. Fig.~\ref{fig:method-b}, Algorithm.~\ref{alg:ZOODiP}, and Algorithm.~\ref{alg:subspace} represents the SG process as described above.

\subsection{Partial Uniform Timestep Sampling}
\label{subsec:PUTS}
To facilitate more efficient training, we introduce Partial Uniform Timestep Sampling (PUTS). The diffusion model can be interpreted as a mixture-of-experts based on the timesteps~\cite{balaji2022ediff,go2024addressing,lee2024multi, liu2023oms}, with each timestep playing a distinct role. In text-to-image diffusion models, several studies~\cite{patashnik2023localizing, choi2022perception, chefer2023attend, zhang2023prospect, daras2022multiresolution} have shown that text conditioning impact varies across timesteps. However, most works focus on inference or training from scratch, while using this insight for personalization remains underexplored.

\begin{figure}[t]
    \centering
    \includegraphics[width=\linewidth]{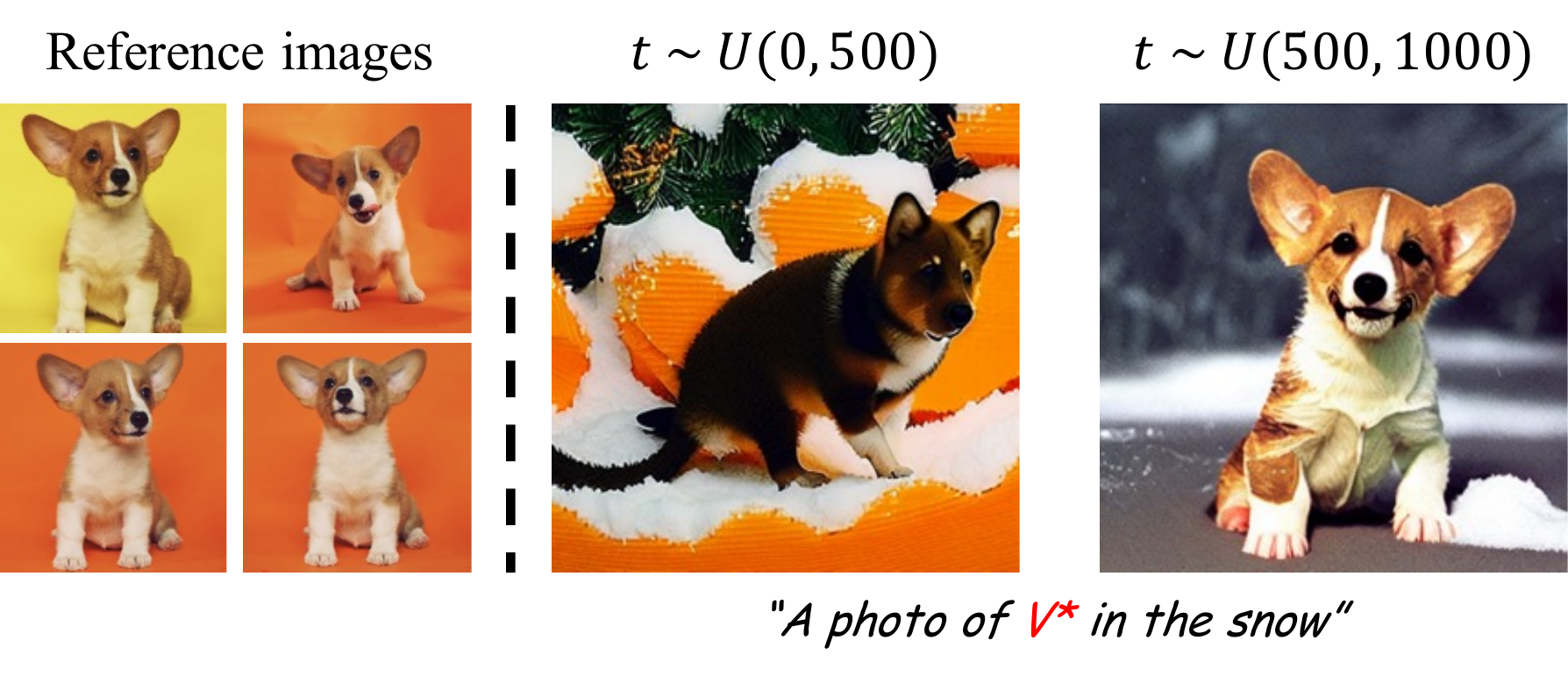}
    \caption{Textual Inversion~\cite{gal2022image} with various timestep sampling. When the timestep $t$ for training is sampled from $U(0,500)$, key conceptual features such as color and body shape of the reference image are not effectively trained. In contrast, sampling from $U(500,1000)$ results in successful learning of these features.}
    \label{fig:timstep}
\end{figure}

TextBoost~\cite{park2024textboost} empirically observed that text influence increases as the timestep nears noise, proposing SNR-based timestep sampling. However, our experiments showed text influence is negligible at certain timesteps (see Fig.~\ref{fig:timstep}). Specifically, sampling timestep $t$ from $U(0, 500)$ failed to effectively learn the reference image concept, whereas sampling from $U(500, 1000)$ led to successful learning.

Based on these observations, we propose PUTS, which focuses on uniformly sampling timesteps within a specific range of the diffusion process. This range is chosen to prioritize timesteps where the text embedding has the most significant influence. PUTS can be formulated as follows:
\begin{equation}
    z_t=\sqrt{\bar{\alpha}_t}z+\sqrt{1-\bar{\alpha}_t}\epsilon, \quad t \sim U\left(T_L, T_U\right)
\end{equation}
where $\epsilon$ represents the noise sampled from Gaussian distribution, $\bar{\alpha}_t$ is the cumulative product of $\alpha$ values to sampled timestep $t$, and $T_{L}$ and $T_{U}$ denote the lower and upper bound of the partial timestep range. Our dense ablation study in Tab.~\ref{tab:ablation_timestep} and Fig.~\ref{fig:PUTS_interval} further display the effectiveness of the proposed method and validate the observations.

\begin{table*}[h!]
    \centering
    \begin{tabular}{cc|cccc|ccc}
        \toprule
        Base. & Method & Quant. & Grad. Free & Mem.$\downarrow$ (GB) & Stor.$\downarrow$ (MB)& CLIP-T$\uparrow$ & CLIP-I$\uparrow$ & DINO$\uparrow$ \\
        \midrule
        \multirow{4}{*}{DB} & DB~\cite{ruiz2023dreambooth} & \redxmark &\redxmark & 19.4 & 3438 &0.281 &  0.782 & 0.592  \\
        & QLoRA~\cite{dettmers2024qlora} & \greencheck &\redxmark  & 7.56 & 1.63 & 0.297 & \underline{0.762}   &  0.607  \\
        & PEQA~\cite{kim2024memory} &\greencheck &\redxmark & 6.31 & 1.32 & \underline{0.275} & 0.791 & 0.604 \\
        & TuneQDM~\cite{ryu2024memory} &\greencheck &\redxmark & 8.96 & 2.48 & 0.289 & 0.788 & \underline{0.555} \\
        \midrule
        \multirow{3}{*}{TI} & TI~\cite{gal2022image} & \redxmark &\redxmark & 6.75 & \textbf{0.003} & 0.285 & 0.778 & 0.559 \\
        & GF-TI~\cite{fei2023gradient} & \greencheck &\greencheck  & \textbf{2.37} & \textbf{0.003} & \doubleunderline{0.253} & \doubleunderline{0.540} & \doubleunderline{0.011}\\    
        & \cellcolor{Yellow}\textbf{ZOODiP (Ours)} & \cellcolor{Yellow}\greencheck &\cellcolor{Yellow}\greencheck  &\cellcolor{Yellow}\textbf{2.37} & \cellcolor{Yellow}\textbf{0.003} & \cellcolor{Yellow}0.287 &\cellcolor{Yellow}0.772 &\cellcolor{Yellow}0.558 \\
        \bottomrule
    \end{tabular}
    \caption{Quantitative comparisons of DreamBooth~\cite{ruiz2023dreambooth} (DB), QLoRA~\cite{dettmers2024qlora} ($r=2$), PEQA~\cite{kim2024memory}, TuneQDM~\cite{ryu2024memory}, Textual Inversion~\cite{gal2022image} (TI), Gradient-Free Textual Inversion~\cite{fei2023gradient} (GF-TI), and Ours. $\uparrow$ / $\downarrow$ indicates higher / lower values are better. Performance was evaluated with CLIP-I and DINO for image alignment, CLIP-T for text-image alignment, and memory requirements of training (Mem.) and storage (Stor.). The \doubleunderline{worst}-performance is double-underlined, and the \underline{second worst} is single-underlined. ZOODiP achieves performance comparable to that of gradient-based methods with significantly less memory.}
    \label{tab:performance}
\end{table*}
\section{Experiments}
\label{sec:experiments}
To evaluate ZOODiP, we conducted quantitative and qualitative comparisons with methods based on DreamBooth (DB)~\cite{ruiz2023dreambooth, kim2024memory, ryu2024memory, dettmers2024qlora} and Textual Inversion (TI)~\cite{gal2022image,fei2023gradient}. Personalization used the DB dataset and Stable Diffusion-v1.5~\cite{rombach2022high}, with INT8 quantization applied to Linear and Conv2D layers. All experiments ran on a single Nvidia RTX 3090 GPU with batch size 1. ZOODiP was trained with  $n=2$, $\mu=10^{-3}$, $\tau=128$, $\nu=10^{-3}$, $T_L=500$, $T_U=900$, $L=30,000$, $\eta=5\times 10^{-3}$ using the ZOAdam~\cite{malladi2023fine} optimizer. Further results and experimental details are provided in the supplementary materials.

\subsection{Quantitative Results}
\paragraph{Image and text alignment score.}
To evaluate ZOODiP's personalization performance, we assessed text and image alignment against baselines. Using the DB dataset, we personalized 30 subjects with 25 prompts, generating 5 images per prompt for a total of 3,750 images. Image alignment was measured via cosine similarity with CLIP~\cite{radford2021learning} (CLIP-I) and DINOv2~\cite{oquab2023dinov2} (DINO) embeddings. Text alignment was measured using cosine similarity between CLIP text embeddings of prompts and generated image embeddings (CLIP-T). Tab.~\ref{tab:performance} shows ZOODiP achieves similar performance to prior methods, outperforming GF-TI by 43.0\% on CLIP-I and 13.4\% on CLIP-T with the same memory usage. Compared to TuneQDM, a state-of-the-art quantized diffusion personalization method, ZOODiP shows a 0.7\% drop on CLIP-T and a 0.5\% increase on DINO. Compared to our baseline, TI, performance differences were +0.7\%, -0.8\%, and -0.2\% on CLIP-T, CLIP-I, and DINO, respectively.

\paragraph{Memory efficiency.}
To assess the training memory efficiency of each method, we tracked peak memory usage using the \texttt{nvidia-smi} command after quantization. As shown in Tab.~\ref{tab:performance}
, ZOODiP requires significantly less memory during training—87.8\% reduction compared to DB and 64.9\% reduction compared to TI, which serves as a baseline for ZOODiP. We also evaluated storage requirements for the optimized models in \texttt{safetensors} format. ZOODiP requires only 3KB of storage, similar to other TI-based methods, making it well-suited for edge devices.

\paragraph{Training speed.}
The training speed of personalization methods is shown in Tab.~\ref{tab:speed}. Since ZOODiP relies solely on forward passes, avoiding costly backpropagation, it achieves significantly faster training speeds compared to other backpropagation-based methods. Specifically, ZOODiP is $2.2\times$ and $1.7\times$ faster than TI for $n=1$ and $n=2$, respectively, and $4.2\times$ and $3.3\times$ faster than TuneQDM. GF-TI is the slowest, with ZOODiP showing a $28\times$ and $22\times$ speed improvement due to GF-TI’s requirement for 30 forward passes per iteration. Additionally, ZOODiP with FP16 precision, fully supported by the hardware, further enhances training speed.

\begin{table}[]
    \centering
    \begin{tabular}{ccc}
        \toprule
        Method & Prec. & Speed (iter/sec) $\uparrow$  \\
        \hline
        TI  & FP32 & 9.42 \\
        \hline
        Ours ($n=1$) & \multirow{2}{*}{FP16} & \textbf{22.3} \\
        Ours ($n=2$) &  & 18.4 \\
        \hline
        TuneQDM &\multirow{4}{*}{INT8}  & 4.94 \\
        GF-TI  &  & 0.74 \\
        Ours ($n=1$) &  & \textbf{20.7} \\
        Ours ($n=2$) &  & 16.1 \\
        \bottomrule
    \end{tabular}
    \caption{Training speed comparisons of prior works and ZOODiP. ZOODiP achieves the fastest training speed by estimating gradients with only forward passes, bypassing backpropagation.}
    \label{tab:speed}
\end{table}


\subsection{Qualitative Results} 
\begin{figure*}
  \centering
  \includegraphics[width=1.0\linewidth]{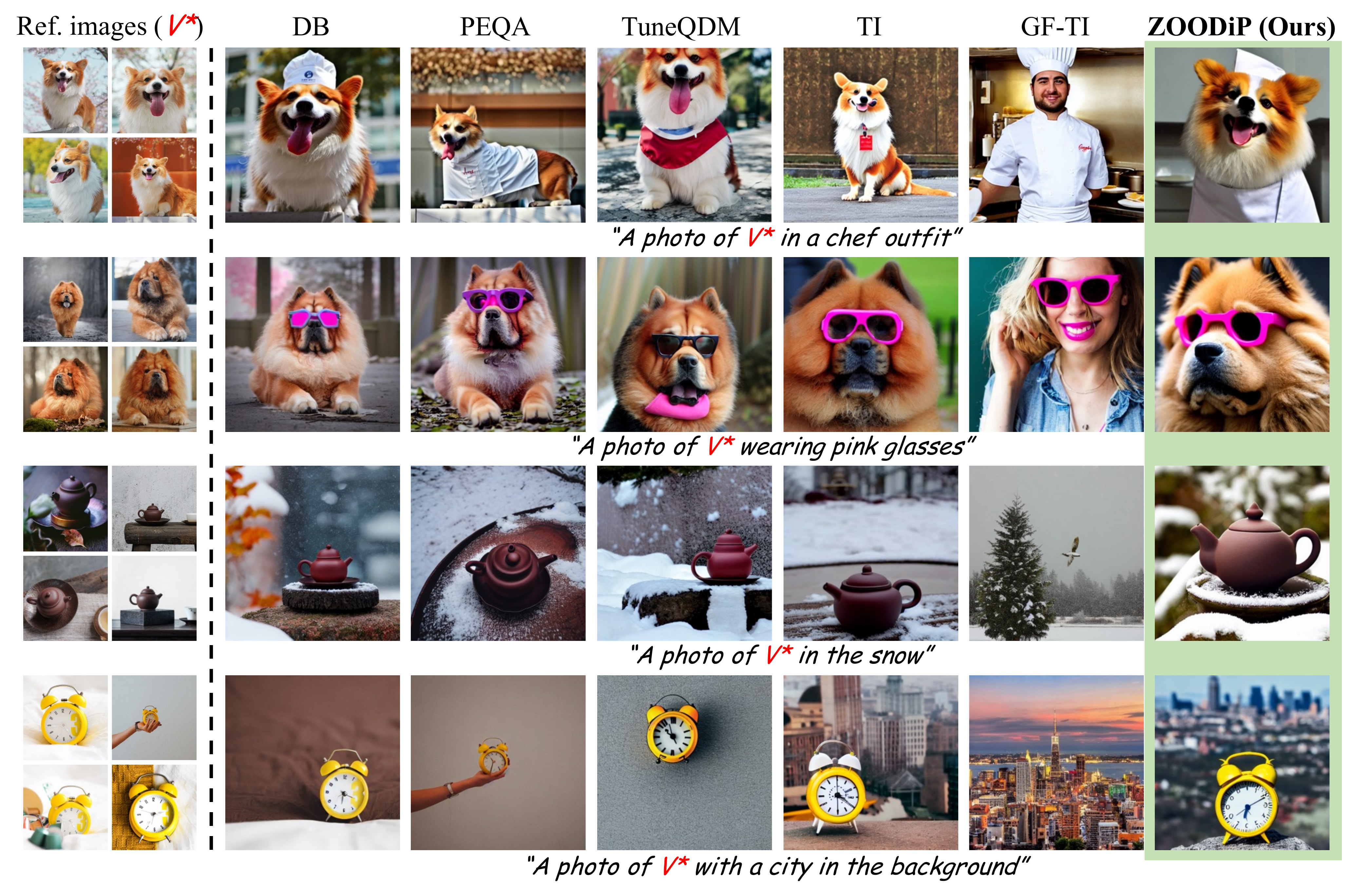}
  \caption{\textbf{Qualitative comparison of image and text alignment.} This figure shows how well each method generates images that match the input text prompt while preserving the identity of the personalized subject. ZOODiP generates images that faithfully reflect the prompt while maintaining the concept of the reference image, demonstrating strong image-text alignment.}
  \label{fig:qualitative_prompt}
\end{figure*}

\begin{figure*}[h]
  \centering
  \includegraphics[width=1.0\linewidth]{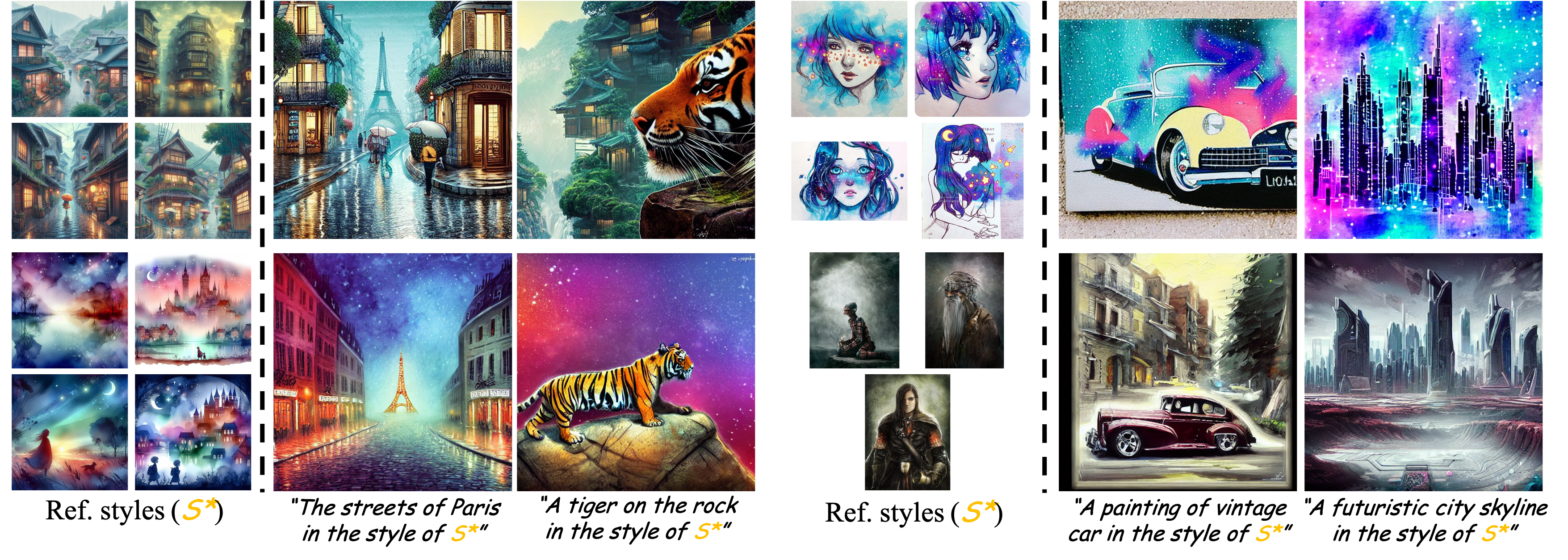}
  \caption{
\textbf{Qualitative results on style personalization.} This figure showcases the results of style personalization achieved through ZOODiP, using few reference images with a consistent style. The outcome highlights ability of ZOODiP to personalize not only the subject but also the style with a high degree of accuracy. This demonstrates the versatility and extensive personalization capabilities of ZOODiP, effectively adapting both stylistic elements and subject details to match the reference images.}
  \label{fig:qualitative_style}
\end{figure*}

Fig.~\ref{fig:qualitative_prompt} shows qualitative comparisons between ZOODiP and other methods. ZOODiP generates images highly faithful to both prompts and reference images, with minimal distinction from other advanced techniques. In contrast, GF-TI, under similar memory constraints, struggles to maintain fidelity, resulting in noticeably inaccurate depictions. 
Fig.~\ref{fig:qualitative_style} shows the results of ZOODiP's style personalization. Using several photos with consistent styles as references, ZOODiP demonstrates accurate reflection of the learned style for the given prompts, indicating that ZOODiP has successfully inherited the style personalization capabilities of TI.

\subsection{Ablation Study}
\label{subsec:ablation}
To analyze the contribution of each component in ZOODiP and assess the impact of hyperparameter variations, we conducted an ablation study. Experiments in this section were performed with \texttt{<dog6>} and \texttt{<shiny\_sneaker>} data from the DreamBooth~\cite{ruiz2023dreambooth} dataset.
\paragraph{Effectiveness of SG and PUTS.} To measure the impact of ZOODiP's key elements—SG and PUTS—we evaluated its performance by incrementally adding each component, as shown in Tab.~\ref{tab:ablation_component}. The results demonstrate that both SG and PUTS significantly enhance performance, outperforming na\"ive Textual Inversion with ZO optimization.

\begin{table}[hbt!]
    \centering
    \begin{tabular}{cc|ccc}
        \toprule
        SG & PUTS& CLIP-T$\uparrow$ & CLIP-I$\uparrow$ & DINO$\uparrow$ \\
        \midrule
         \xmark & \xmark & 0.273 & 0.736 & 0.505 \\
         \cmark & \xmark & 0.265 & 0.747 & 0.527 \\
         \xmark & \cmark & \textbf{0.277} & 0.744 & 0.562 \\
         \cmark & \cmark & 0.266 & \textbf{0.759} & \textbf{0.569} \\
        \bottomrule
    \end{tabular}
    \caption{Ablation study on ZOODiP components with \texttt{<shiny\_sneaker>}. \cmark~denotes the component is applied, while \xmark~means it is not. Without PUTS, timesteps are sampled uniformly.}
    \label{tab:ablation_component}
\end{table}

\paragraph{Hyperparameter study.}
We conducted exhaustive experiments to analyze the impact of two key hyperparameters used in SG: $\tau$ and $\nu$. As shown in Tab.~\ref{tab:ablation_hyperparam}, ZOODiP shows robust performance unless $\tau$ or $\nu$ is too large or small. When $\tau$ is small, frequent $P_{\nu}$ updates may occur, introduce computational overhead, so we choose values of $\tau=128$ and $\nu=10^{-3}$ which show the best performance.

\begin{table}[t]
    \centering
    \begin{tabular}{c|cccc}
        \toprule
        \diagbox[width=3em]{$\tau$}{$\nu$} & $10^{-1}$ & $10^{-2}$ & $10^{-3}$ & $10^{-4}$ \\
        \midrule
        32  & 0.704  & 0.686  & 0.716  & 0.729 \\
        64  & 0.721  & 0.736  & 0.712  & 0.724 \\
        128 & 0.736  & 0.735  & \textbf{0.759} & 0.716 \\
        256 & 0.739  & 0.727  & 0.738  & 0.716 \\
        512 & 0.705  & 0.704  & 0.676  & 0.707 \\
        \bottomrule
    \end{tabular}
    
    \caption{Ablation study on hyperparameters $\tau$ and $\nu$ incorporated with SG. We optimized the pseudo-token with various $\tau$ and $\nu$ and measured the performance with the CLIP-I score. Experiments were done with \texttt{<shiny\_sneaker>} dataset.}
    \label{tab:ablation_hyperparam}
\end{table}

\paragraph{Diffusion timestep sampling.}
To assess the effectiveness of PUTS on performance, we conducted experiments with different diffusion timestep sampling strategies: uniform sampling, SNR-based sampling~\cite{park2024textboost}, and PUTS. The results in Tab.~\ref{tab:ablation_timestep} confirm that PUTS outperforms the other sampling methods, validating its effectiveness.

\begin{figure}[hbt!]
  \centering
    \begin{subfigure}{0.32\linewidth}
    \includegraphics[width=1.0\linewidth]{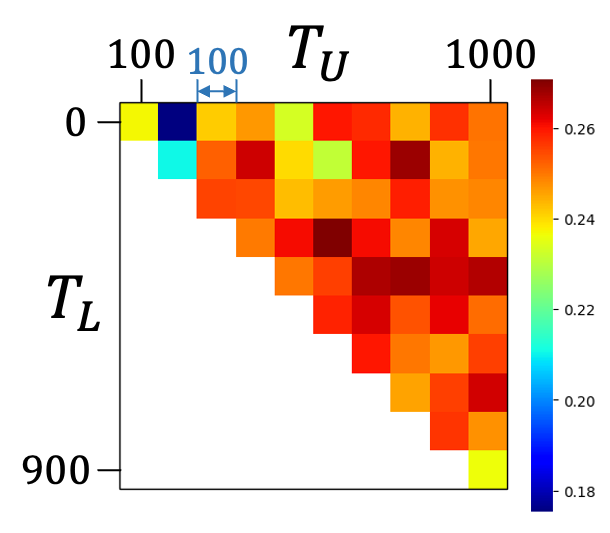}
    \caption{CLIP-T scores.}
    \label{fig:PUTS_clip_t}
  \end{subfigure}
  \begin{subfigure}{0.32\linewidth}
    \includegraphics[width=1.0\linewidth]{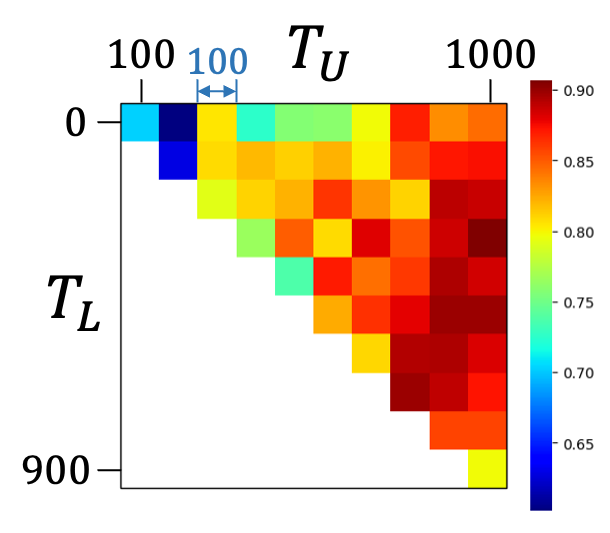}
    \caption{CLIP-I scores.}
    \label{fig:PUTS_clip_i}
  \end{subfigure}
  \begin{subfigure}{0.32\linewidth}
    \includegraphics[width=1.0\linewidth]{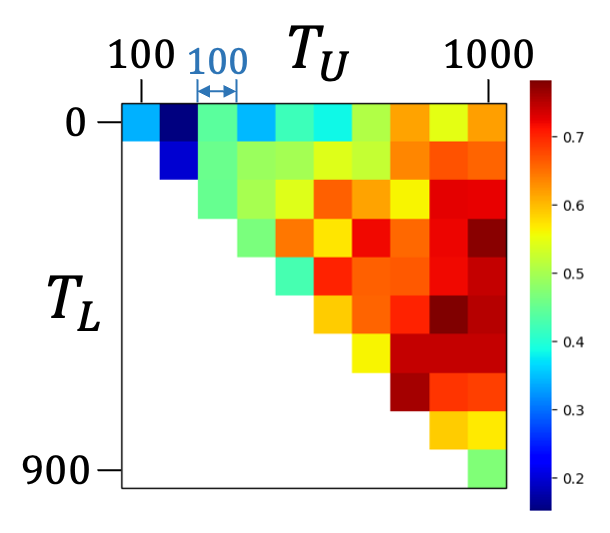}
    \caption{DINO scores.}
    \label{fig:PUTS_dino}
  \end{subfigure}
  \caption{Heatmap of CLIP-T, CLIP-I and DINO scores across varying $T_L$ and $T_U$ on the \texttt{<dog6>} dataset. $x$-axis is the $T_U$ and $y$-axis is the $T_L$ applied to the sampling distribution.}
  \label{fig:PUTS_interval}
\end{figure}

Furthermore, we conducted exhaustive experiments on various combinations of  $T_U$ and $T_L$. As shown in Fig.~\ref{fig:PUTS_interval}, we divided the diffusion timesteps into 10 units and tested all possible combinations. The results indicate that training with timesteps closer to the noise improves both image and text fidelity, supporting our choice of $T_L$ and $T_U$ values.

\begin{table}[t]
    \centering
    \begin{tabular}{c|cccc}
        \toprule
        Method & CLIP-T$\uparrow$ & CLIP-I$\uparrow$ & DINO$\uparrow$  \\
        \midrule
        Uniform              & 0.265 & 0.747 & 0.527 \\
        SNR-based            & \textbf{0.271} & 0.719 & 0.545 \\
        \textbf{PUTS (Ours)} & 0.266 & \textbf{0.759} & \textbf{0.569} \\
        \bottomrule
    \end{tabular}
    \caption{Ablation study on various diffusion timestep sampling method with \texttt{<shiny\_sneaker>}. PUTS outperforms in image alignment score among all sample methods with minor degradation in text alignment score compared to SNR-based sampling.}
    \label{tab:ablation_timestep}
\end{table}

\section{Discussion}
\label{sec:discussion}
\begin{figure}
  \centering
  \includegraphics[width=\linewidth]{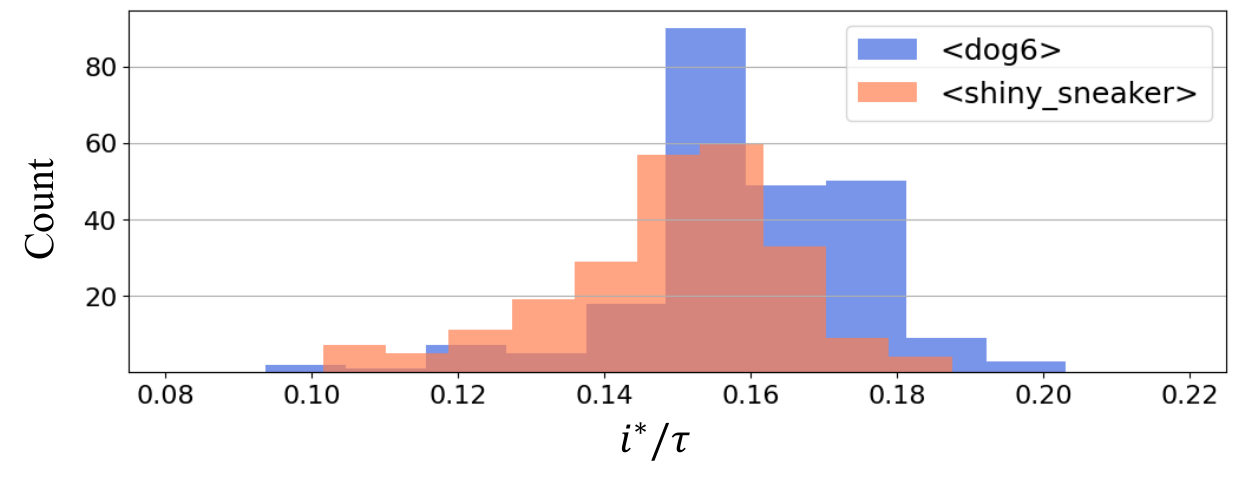}
  \caption{Histogram of $i^*/\tau$ ratios for \texttt{<dog6>} and \texttt{<shiny\_sneaker>} dataset with hyperparameter $\tau=128$, $\nu=10^{-3}$ during training with SG. Despite the small $\nu$, a significant portion ($>80\%$) of dimensions are projected out.}
  \label{fig:discussion_ratio}
\end{figure}

ZO optimization's convergence rate is influenced by the effective dimension of the optimized parameters~\cite{yue2023zeroth, malladi2023fine, ling2024convergence}. We conjecture that ZOODiP's success arises from the low dimensionality of the optimized token and its even smaller effective dimension (Sec.~\ref{subsec:SG}). The concentrated variance of the token trajectory allows SG to effectively project out over two-thirds of the dimensions in the trajectory buffer, even with a small $\nu$ value. It aligns with Fig.~\ref{fig:subspace} and the analysis in Sec.~\ref{subsec:SG}. We selected $\nu=10^{-3}$ to concentrate variance ratios within one-third of the total (Fig.~\ref{fig:discussion_ratio}).
\section{Conclusion}
\label{sec:conclusion}
We propose ZOODiP, a framework that enables the personalization of diffusion models even in highly memory-constrained environments. ZOODiP leverages quantized models during training to reduce memory footprint. We then propose zeroth-order optimization to personalize these models. To overcome the limitations of zeroth-order optimization, we introduce Subspace Gradient (SG), which utilizes the trajectory of the optimizing token to eliminate noisy gradient dimensions and enhance performance based on the empirical analysis of the effective dimension of personalized tokens. Additionally, we systematically analyzed the impact of diffusion timesteps on personalization and identified the most effective timesteps for efficient training. Based on this analysis, we proposed Partial Uniform Timestep Sampling (PUTS), which samples only these relevant timesteps. Through extensive qualitative and quantitative experiments, we demonstrate that ZOODiP achieves performance comparable to prior arts while requiring significantly less memory of only 2.37GB during training.

\section*{Acknowledgements}
This work was supported in part by Institute of Information \& communications Technology Planning \& Evaluation (IITP) grant funded by the Korea government(MSIT) [NO. RS-2021-II211343, Artificial Intelligence Graduate School Program (Seoul National University)], National Research Foundation of Korea (NRF) grant funded by the Korea government (MSIT) (No. NRF-2022M3C1A309202211) and Samsung Electronics MX Division. Also, the authors acknowledged the financial support from the BK21 FOUR program of the Education and Research Program for Future ICT Pioneers, Seoul National University.

\renewcommand{\thefigure}{S\arabic{figure}}
\renewcommand{\thetable}{S\arabic{table}}
\renewcommand{\thesection}{S\arabic{section}}
\renewcommand{\theequation}{S\arabic{equation}}

\setcounter{section}{0}
\setcounter{figure}{0}
\setcounter{table}{0}
\section*{Appendix}

\section{Additional Experimental Details}
\subsection{Dataset}
All quantitative experiments were conducted using the DreamBooth~\cite{ruiz2023dreambooth} (DB) dataset, which includes 30 distinct subjects, which are denoted in Tab.~\ref{tab:supp_subject_name}, each paired with 25 unique prompts. Among these subjects, nine are living entities—specifically dogs and cats—while the remaining 21 represent non-living objects. Details of the subjects and their respective prompts are provided in the Tab.~\ref{tab:supp_eval_living} (living) and Tab.~\ref{tab:supp_eval_non_living} (non-living). Notably, the Textual Inversion~\cite{gal2022image}-based method did not incorporate class tokens in the prompts used for image generation, ensuring a fair comparison by excluding scenarios where class tokens are utilized during the unique token learning process. For Textual Inversion~\cite{gal2022image} (TI) and Gradient-Free Textual Inversion~\cite{fei2023gradient} (GF-TI), we employed the ImageNet-based template proposed in the original TI paper for training. ZOODiP followed the DCO~\cite{lee2024direct}, utilizing Comprehensive Captioning (CC) for the text prompt.
{
\begin{table*}[!h]
\centering
{\footnotesize
\begin{tabular}{p{\dimexpr \linewidth-2\tabcolsep\relax}} 
\hline
\textbf{Subjects in DreamBooth~\cite{ruiz2023dreambooth} dataset} \\ \hline
backpack, backpack\_dog, bear\_plushie, berry\_bowl, can, candle, cat, cat2, clock, colorful\_sneaker, dog, dog2, dog3, dog5, dog6, dog7, dog8, duck\_toy, fancy\_boot, grey\_sloth\_plushie, monster\_toy, pink\_sunglasses, poop\_emoji, rc\_car, red\_cartoon, robot\_toy, shiny\_sneaker, teapot, vase, wolf\_plushie \\ \hline
\end{tabular}
}
\vspace{-0.5em}
\caption{Full subjects name of DreamBooth dataset. 
The dataset names in the main paper are based on the corresponding subject datasets.}
\label{tab:supp_subject_name}
\end{table*}

\begin{table*}[!h]
\centering
{\footnotesize
\begin{tabular}{p{\dimexpr 0.01\linewidth}
                 p{\dimexpr 0.93\linewidth}}
\hline
\multicolumn{2}{l}{\textbf{Full prompt used in evaluation (living)}}                                                                                      \\ \hline
1  & `\texttt{a \{0\} \{1\} in the jungle'.format(unique\_token, class\_token)} \\
2  & `\texttt{a \{0\} \{1\} in the snow'.format(unique\_token, class\_token)} \\
3  & `\texttt{a \{0\} \{1\} on the beach'.format(unique\_token, class\_token)} \\
4  & `\texttt{a \{0\} \{1\} on a cobblestone street'.format(unique\_token, class\_token)} \\
5  &`\texttt{a \{0\} \{1\} on top of pink fabric'.format(unique\_token, class\_token)} \\
6  &`\texttt{a \{0\} \{1\} on top of a wooden floor'.format(unique\_token, class\_token)} \\
7  &`\texttt{a \{0\} \{1\} with a city in the background'.format(unique\_token, class\_token)} \\
8  & `\texttt{a \{0\} \{1\} with a mountain in the background'.format(unique\_token, class\_token)} \\
9  & `\texttt{a \{0\} \{1\} with a blue house in the background'.format(unique\_token, class\_token)} \\
10 & `\texttt{a \{0\} \{1\} on top of a purple rug in a forest'.format(unique\_token, class\_token)} \\
11 & `\texttt{a \{0\} \{1\} wearing a red hat'.format(unique\_token, class\_token)} \\
12 & `\texttt{a \{0\} \{1\} wearing a santa hat'.format(unique\_token, class\_token)} \\
13 & `\texttt{a \{0\} \{1\} wearing a rainbow scarf'.format(unique\_token, class\_token)} \\
14 & `\texttt{a \{0\} \{1\} wearing a black top hat and a monocle'.format(unique\_token, class\_token)} \\
15 & `\texttt{a \{0\} \{1\} in a chef outfit'.format(unique\_token, class\_token)} \\
16 & `\texttt{a \{0\} \{1\} in a firefighter outfit'.format(unique\_token, class\_token)} \\
17 & `\texttt{a \{0\} \{1\} in a police outfit'.format(unique\_token, class\_token)} \\
18 & `\texttt{a \{0\} \{1\} wearing pink glasses'.format(unique\_token, class\_token)} \\
19 & `\texttt{a \{0\} \{1\} wearing a yellow shirt'.format(unique\_token, class\_token)} \\
20 & `\texttt{a \{0\} \{1\} in a purple wizard outfit'.format(unique\_token, class\_token)} \\
21 & `\texttt{a red \{0\} \{1\}'.format(unique\_token, class\_token)} \\
22 & `\texttt{a purple \{0\} \{1\}'.format(unique\_token, class\_token)} \\
23 & `\texttt{a shiny \{0\} \{1\}'.format(unique\_token, class\_token)} \\
24 & `\texttt{a wet \{0\} \{1\}'.format(unique\_token, class\_token)} \\
25 & `\texttt{a cube shaped \{0\} \{1\}'.format(unique\_token, class\_token)} \\
\hline
\end{tabular}
}
\vspace{-0.5em}
\caption{Full prompts used in evaluation of living category objects. \texttt{unique\_token} represents the special token corresponds to object which aims to personalize, and \texttt{class\_token} denotes the class that \texttt{unique\_token} is in.}
\label{tab:supp_eval_living}
\end{table*}

\begin{table*}[!h]
\centering
{\footnotesize
\begin{tabular}{ p{\dimexpr 0.01\linewidth}
                 p{\dimexpr 0.93\linewidth}}
\hline
\multicolumn{2}{l}{\textbf{Full prompt used in evaluation (non-living)}}                                                                                      \\ \hline
1  & `\texttt{a \{0\} \{1\} in the jungle'.format(unique\_token, class\_token)} \\
2 & `\texttt{a \{0\} \{1\} in the snow'.format(unique\_token, class\_token)} \\
3 & `\texttt{a \{0\} \{1\} on the beach'.format(unique\_token, class\_token)} \\
4 & `\texttt{a \{0\} \{1\} on a cobblestone street'.format(unique\_token, class\_token)} \\
5 &`\texttt{a \{0\} \{1\} on top of pink fabric'.format(unique\_token, class\_token)} \\
6&`\texttt{a \{0\} \{1\} on top of a wooden floor'.format(unique\_token, class\_token)} \\
7&`\texttt{a \{0\} \{1\} with a city in the background'.format(unique\_token, class\_token)} \\
8 & `\texttt{a \{0\} \{1\} with a mountain in the background'.format(unique\_token, class\_token)} \\
9 & `\texttt{a \{0\} \{1\} with a blue house in the background'.format(unique\_token, class\_token)} \\
10 & `\texttt{a \{0\} \{1\} on top of a purple rug in a forest'.format(unique\_token, class\_token)} \\
11 & `\texttt{a \{0\} \{1\} with a wheat field in the background'.format(unique\_token, class\_token)} \\
12 & `\texttt{a \{0\} \{1\} with a tree and autumn leaves in the background'.format(unique\_token, class\_token)} \\
13 & `\texttt{a \{0\} \{1\} with the Eiffel Tower in the background'.format(unique\_token, class\_token)} \\
14 & `\texttt{a \{0\} \{1\} floating on top of water.format(unique\_token, class\_token)} \\
15 & `\texttt{a \{0\} \{1\} floating in an ocean of milk'.format(unique\_token, class\_token)} \\
16 & `\texttt{a \{0\} \{1\} on top of green grass with sunflowers around it'.format(unique\_token, class\_token)} \\
17 & `\texttt{a \{0\} \{1\} on top of a mirror'.format(unique\_token, class\_token)} \\
18 & `\texttt{a \{0\} \{1\} on top of the sidewalk in a crowded street'.format(unique\_token, class\_token)} \\
19 & `\texttt{a \{0\} \{1\} on top of a dirt road'.format(unique\_token, class\_token)} \\
20 & `\texttt{a \{0\} \{1\} on top of a white rug'.format(unique\_token, class\_token)} \\
21 & `\texttt{a red \{0\} \{1\}'.format(unique\_token, class\_token)} \\
22 & `\texttt{a purple \{0\} \{1\}'.format(unique\_token, class\_token)} \\
23 & `\texttt{a shiny \{0\} \{1\}'.format(unique\_token, class\_token)} \\
24 & `\texttt{a wet \{0\} \{1\}'.format(unique\_token, class\_token)} \\
25 & `\texttt{a cube shaped \{0\} \{1\}'.format(unique\_token, class\_token)} \\
\hline
\end{tabular}
}
\vspace{-0.5em}
\caption{Full prompts used in evaluation of non-living category objects. \texttt{unique\_token} represents the special token corresponds to object which aims to personalize, and \texttt{class\_token} denotes the class that \texttt{unique\_token} is in.}
\label{tab:supp_eval_non_living}
\end{table*}
}

\subsection{Metrics}
To compute the CLIP-I score, we use the \texttt{openai/clip-vit-base-patch32} model~\cite{radford2021learning} from Huggingface to extract image features for both the reference and generated images. We then calculate the cosine similarity for all possible pairs and average these values. For the CLIP-T score, we leverage the same model's text encoder to extract text features from the input prompt. We also calculate image features for the generated images and measure the pairwise cosine similarity between the text and image features, averaging the results to obtain the final score. To compute the DINO score, we utilize the \texttt{facebook/dinov2-base} model from Huggingface to extract DINOv2~\cite{oquab2023dinov2} embeddings for both the reference and generated images. The score is determined by averaging the pairwise cosine similarities between these embeddings.

\subsection{Quantization}
We utilized \texttt{optimum-quanto}, the PyTorch~\cite{paszke2019pytorch} quantization backend provided by Huggingface, to enable accelerated matrix multiplications on CUDA devices. For the 8-bit weight quantization described in the main paper, we employed \texttt{optimum-quanto} to quantize the weights of all Linear and Conv2D layers in the VAE, U-Net, and text encoder modules to 8-bit integer. Other layers' parameters including text embedding, layer normalization, and batch normalization were not quantized. With this process, 96.4\% of the whole Stable Diffusion~\cite{rombach2022high} pipeline (U-Net, VAE, and text encoder) parameters are quantized to INT8, while the other 3.6\% parameters remain FP16.

\subsection{Memory usage}
We primarily monitored memory usage using the \texttt{nvidia-smi} command to observe real-time memory consumption during training and employed the PyTorch profiler for detailed memory breakdowns. Notably, the memory utilized by the CUDA context can vary based on the CUDA and PyTorch versions, so the reported memory usage may differ depending on the experimental environment. Our measurements were conducted under the following settings: CUDA Toolkit 11.8, Torch 2.4.1, Torchvision 0.19.1, and Python 3.8.10. Since PyTorch loads all CUDA libraries by default, including unused ones, the actual memory usage could potentially be reduced by unloading unnecessary libraries. However, as our focus was not on such optimizations, we ensured a fair comparison by evaluating all training methods under the same experimental environment. We also observed that Variational Auto-Encoder (VAE) encoding introduces significant memory overhead. Given that all methods do not require backpropagation through the VAE, we standardized the process by blocking gradients in the VAE encoding step across all methods. This adjustment ensured that each personalization method used the minimal memory required, facilitating a fair evaluation of memory usage.

\subsection{Baseline training configuration}
\label{sec:supp_config}
For our experiments, we utilized the unofficial implementations provided by Huggingface for Textual Inversion~\cite{gal2022image} and DreamBooth~\cite{ruiz2023dreambooth} without prior-preservation loss which is suitable to memory-constrained environment. For PEQA~\cite{kim2024memory} and TuneQDM~\cite{ryu2024memory}, we conducted experiments using reproduced code based on the Huggingface implementation of DreamBooth. QLoRA~\cite{dettmers2024qlora} was implemented using the \texttt{BitsandBytes} library. For Gradient-Free Textual Inversion~\cite{fei2023gradient}, we utilized the official implementation for our experiments. We configured the training settings for each method to be as consistent as possible, adjusting only the batch size to 1. For DreamBooth, we set $\eta=5\times10^{-6}$, $L=400$, for Textual Inversion, we set $\eta=5\times10^{-3}$, $L=5,000$, for QLoRA, we set $\eta=1\times10^{-4}$, $L=500$, for PEQA, we set $\eta=3\times10^{-6}$, $L=400$, for TuneQDM, we set $\eta=3\times10^{-6}$, $L=400$, for Gradient-Free Textual Inversion, $\eta=5\times10^{6}$, $L=13,000$, intrinsic dimension $d_i=256$, $\sigma=1$, and $\alpha=1$.

\section{Additional Ablation Study}
We conducted an extensive ablation study to analyze the impact of each component of ZOODiP. Beyond evaluating RGE, we examined the effectiveness of alternative gradient estimation methods, the influence of varying perturbation scales on performance, the impact of changing the number of gradient estimations, and the resulting time overhead associated with these changes. Unless otherwise noted, all additional experiments were performed on two subjects: \texttt{<dog6>} as a representative of living objects and \texttt{<shiny\_sneaker>} as a representative of non-living objects. The results from these experiments were averaged to ensure a balanced and comprehensive evaluation.

\begin{table}[]
    \centering
    \begin{tabular}{c|ccc}
    \toprule
        Method    & CLIP-T$\uparrow$ & CLIP-I$\uparrow$ & DINO$\uparrow$ \\
    \midrule
        RGE       & 0.266 & 0.759 & 0.569  \\
        SPSA      & 0.277 & 0.732 & 0.506  \\
        One-point & 0.296 & 0.703 & 0.393  \\
    \bottomrule
    \end{tabular}
    \caption{Personalization performance across different gradient estimation methods with $n=2$. The results show that RGE outperforms other two different gradient estimations methods. Notably, RGE is more efficient in terms of computational cost, requiring fewer forward passes than SPSA when $n > 1$. Due to this efficiency and performance advantages, we adopted RGE as the gradient estimation method in ZOODiP.}
    \label{tab:supp_gradient_estimation}
\end{table}

\subsection{Gradient estimation method}
There are various methods for estimating gradients, each with unique advantages and trade-offs. In ZOODiP, we opted for Random Gradient Estimation (RGE, Eq.~\ref{eq:supp_RGE}) due to its efficiency in estimating gradients using Monte Carlo gradient estimation. While many existing works utilize Simultaneous Perturbation Stochastic Approximation (SPSA, Eq.~\ref{eq:supp_SPSA}), which perturbs the gradient in two directions, or a one-point estimation method (Eq.~\ref{eq:supp_one_point}) that calculates the gradient directly at the perturbed point, we evaluated all three methods for gradient estimation as seen in Tab.~\ref{tab:supp_gradient_estimation}.

\begin{equation}\label{eq:supp_RGE}
\hat{g}_{\theta} = \frac{1}{n}\sum_{i=1}^{n}\Bigg[ \frac{\mathcal{L}_{\text{LDM}}(\theta + \mu e_i) - \mathcal{L}_{\text{LDM}}(\theta)}{\mu}e_i \Bigg]
\end{equation}
\begin{equation}\label{eq:supp_SPSA}
\hat{g}_{\theta} = \frac{1}{n}\sum_{i=1}^{n}\Bigg[ \frac{\mathcal{L}_{\text{LDM}}(\theta + \mu e_i) - \mathcal{L}_{\text{LDM}}(\theta - \mu e_i)}{2\mu}e_i \Bigg]
\end{equation}
\begin{equation}\label{eq:supp_one_point}
\hat{g}_{\theta} = \frac{1}{n}\sum_{i=1}^{n}\Bigg[ \frac{\mathcal{L}_{\text{LDM}}(\theta + \mu e_{i+1}) - \mathcal{L}_{\text{LDM}}(\theta + \mu e_i)}{\mu}e_i \Bigg]
\end{equation}

Our results indicate that SPSA and RGE exhibit similar performance, with RGE occasionally outperforming SPSA. In contrast, the one-point estimation method performed poorly. From a computational perspective, SPSA requires $2n$ forward passes for $n$ gradient estimation steps, whereas RGE needs only $n+1$ forward passes, making it more efficient. Given its favorable balance of computational efficiency and performance, we selected RGE as the gradient estimation method for ZOODiP.

\begin{table}[t]
    \centering
    \begin{tabular}{c|cccc}
        \toprule
        $\mu$ & CLIP-T$\uparrow$ & CLIP-I$\uparrow$ & DINO$\uparrow$  \\
        \midrule
        $1\times10^{-2}$ & 0.281 & 0.778 & 0.599 \\
        $\mathbf{1\times10^{-3}}$ & 0.277 & \textbf{0.797} & \textbf{0.613}  \\
        $1\times10^{-4}$ & \textbf{0.296} & 0.724 & 0.470 \\
        \bottomrule
    \end{tabular}
    \caption{Quantitative results for different $\mu$ values. Optimal performance is observed at $\mu=10^{-3}$ with varying $\mu$, and we have set the value of $\mu$ accordingly for ZOODiP.}
    \label{tab:supp_perturb_scale}
\end{table}

\subsection{Perturbation scale}
We analyzed the impact of varying $\mu$, the scaling parameter (\textit{a.k.a} smoothing parameter) for perturbations in Eq.~\ref{eq:supp_RGE}, on personalization performance. In general, the optimal value of $\mu$ can vary depending on the configuration of the model to be tuned as seen in Sec.~\ref{sec:supp_4bit} and Sec.~\ref{supp_sec:general}. The results in Tab.~\ref{tab:supp_perturb_scale} demonstrate that the chosen value of $10^{-3}$ is the most suitable perturbation scale for our setup.

\begin{table}[t]
    \centering
    \begin{tabular}{c|ccc|cc}
        \toprule
        $n$ & CLIP-T$\uparrow$ & CLIP-I$\uparrow$ & DINO$\uparrow$ & Speed$\uparrow$ (iter/s)  \\
        \midrule
        1 & 0.298 & 0.736 & 0.495 & 20.7 \\
        2 & 0.277 & 0.796 & 0.613 & 16.1 \\
        4 & 0.282 & 0.784 & 0.584 & 9.78 \\
        8 & 0.282 & 0.798 & 0.627 & 6.20 \\
        \bottomrule
    \end{tabular}
    \caption{Quantitative results for various $n$, the number of gradient estimations. $n=2$ is the promising choice between performance and computation efficiency.}
    \label{tab:supp_grad_estimate_num}
\end{table}

\subsection{Number of gradient estimation}
According to MeZO~\cite{malladi2023fine}, increasing $n$ when fine-tuning large language models provides only marginal performance gains compared to the proportional increase in the number of forward passes, making $n=1$ the most efficient choice. Tab.~\ref{tab:supp_grad_estimate_num} illustrates the personalization performance across different $n$ values with RGE. While increasing $n$ slightly improves performance due to more accurate gradient estimation, the associated increase in training time becomes a limiting factor. Thus, we set $n=2$ as a compromise between performance and efficiency.

\begin{table}[t]
    \centering
    \begin{tabular}{c|c|ccc}
        \toprule
        Method & PUTS & CLIP-T$\uparrow$ & CLIP-I$\uparrow$ & DINO$\uparrow$ \\
        \midrule
        \multirow{2}{*}{DB} & \redxmark & 0.266  & 0.853  & 0.751   \\
        & \greencheck & \textbf{0.272} & \textbf{0.854} & \textbf{0.758}  \\
        \midrule
        \multirow{2}{*}{TI} & \redxmark & 0.241 & 0.798 & 0.584  \\
        & \greencheck & \textbf{0.247} & \textbf{0.825} & \textbf{0.661}  \\
        \bottomrule
    \end{tabular}
    \caption{Quantitative results from applying PUTS to DreamBooth and Textual Inversion. The results confirm that PUTS enhances the performance of gradient-based personalization methods by up to 13.2\%. The improvement was particularly pronounced in Textual Inversion, which is highly influenced by the text encoder, highlighting the significant impact of PUTS in this context.}
    \label{tab:supp_fo_puts}
\end{table}

\section{Additional Experiments}
\subsection{PUTS on gradient-based methods}
Partial Uniform Timestep Sampling (PUTS), one of the key components of ZOODiP, not only enhances ZOODiP's performance but also proves effective in gradient-based approaches. To validate this, we conducted experiments using two representative gradient-based personalization methods on Stable Diffusion v1.5: Textual Inversion and DreamBooth. These experiments were performed with the same hyperparameter settings for $T_L$ and $T_U$ as those used in ZOODiP. The results in Tab.~\ref{tab:supp_fo_puts} demonstrate that applying Partial Uniform Timestep Sampling (PUTS) to gradient-based methods improves performance by up to 11.6\%. PUTS enhances efficiency by prioritizing informative timesteps, showcasing its versatility in optimizing both zeroth-order and gradient-based approaches.

\begin{table}[t]
    \centering
    \begin{tabular}{c|ccc}
        \toprule
        U-Net Precision & CLIP-T$\uparrow$ & CLIP-I$\uparrow$ & DINO$\uparrow$  \\
        \midrule
        INT8 & \textbf{0.288} & 0.834 & \textbf{0.657} \\
        INT4 & 0.212 & \textbf{0.835} & 0.647 \\
        \bottomrule
    \end{tabular}
    \caption{The performance comparison between INT8 and INT4 for U-Net precision on the \texttt{<dog6>} subset of the DB dataset.}
    \label{tab:supp_4bit}
\end{table}

\begin{figure}
  \centering
  \includegraphics[width=\linewidth]{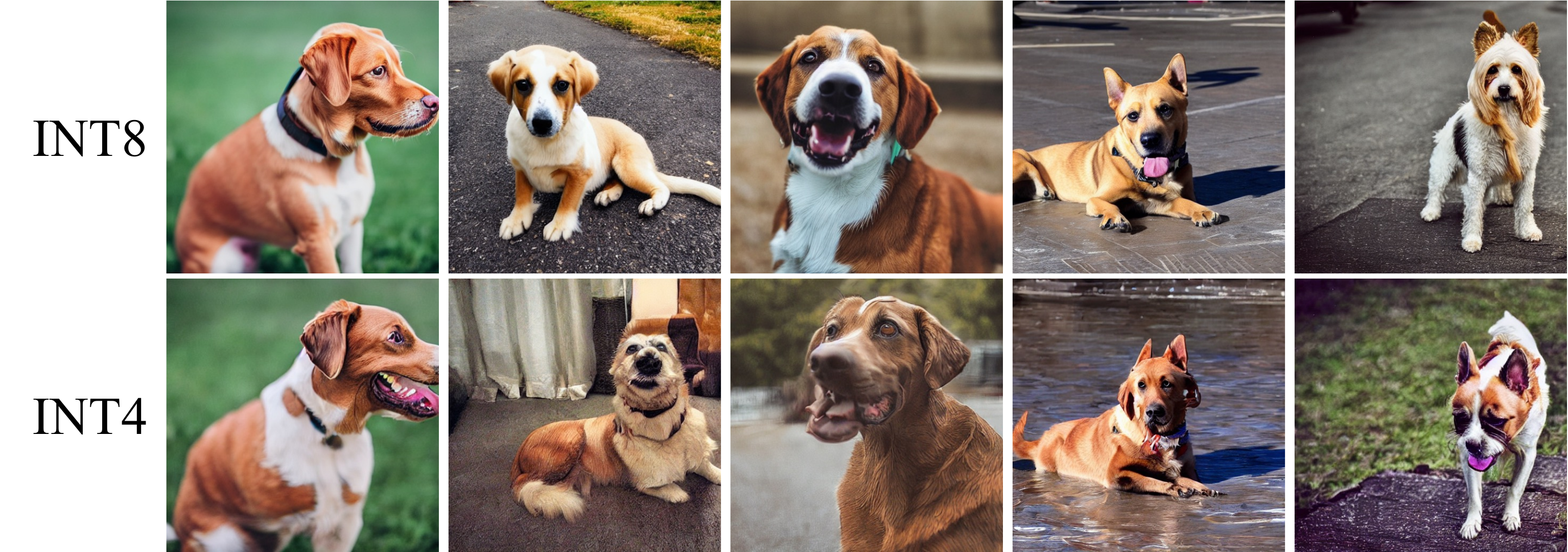}
  \caption{Generated images with the prompt \textit{``a photo of a dog''} with various weight precision. While INT8 precision produces results nearly equivalent to full-precision performance, INT4 precision exhibits noticeable degradation in image quality, highlighting the trade-off between lower precision and fidelity.}
  \label{fig:supp_int4_deg}
\end{figure}

\subsection{4-bit quantized models}
\label{sec:supp_4bit}
\begin{figure}
  \centering
  \includegraphics[width=\linewidth]{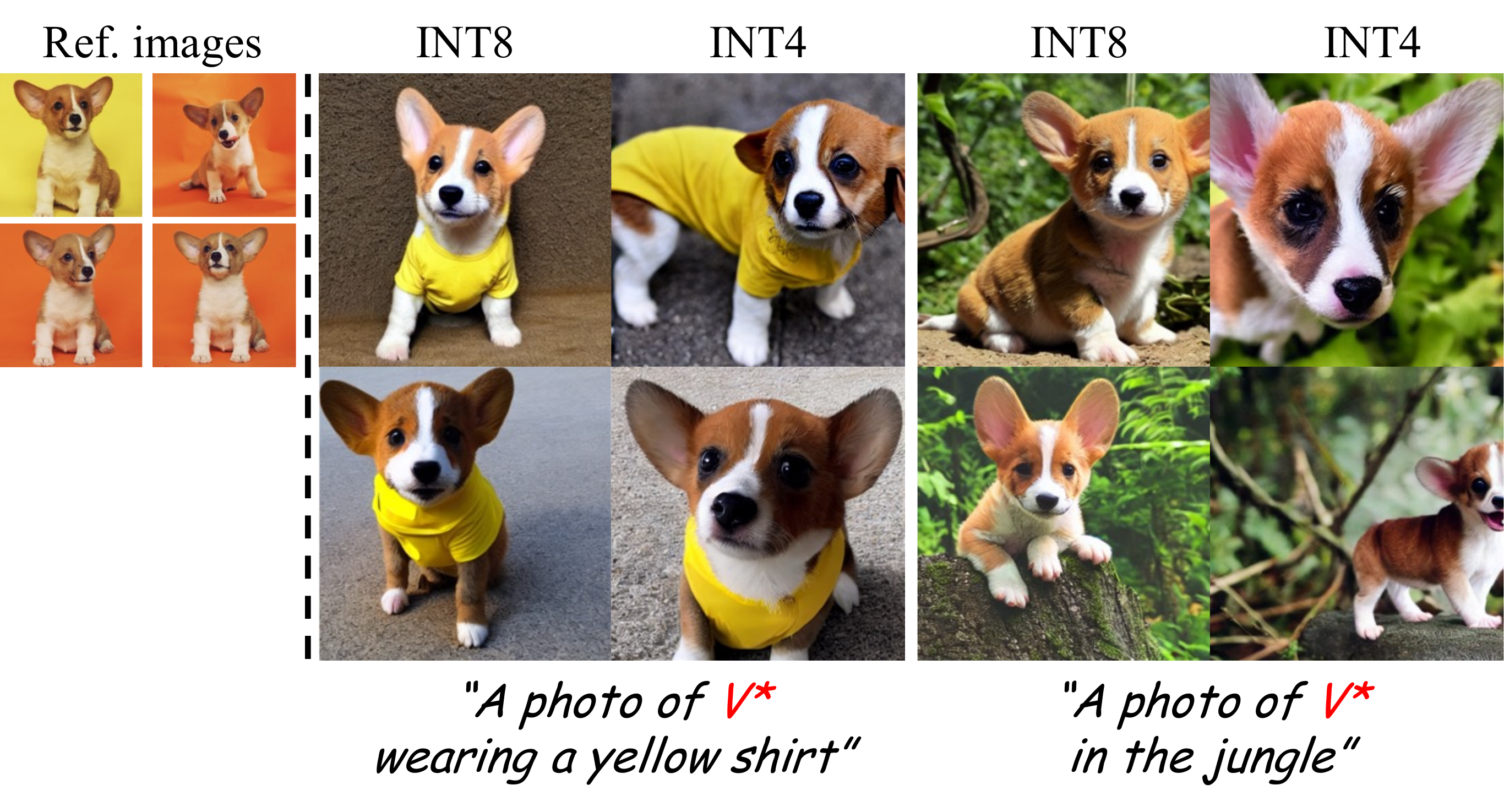}
  \caption{Qualitative results of U-Net precision at INT8 and INT4 in \texttt{<dog6>} dataset. ZOODiP works on INT4 and INT8, but performance diminishes due to degradation caused by INT4 quantization.}
  \label{fig:supp_int4}
\end{figure}

We applied ZOODiP to a 4-bit quantized model of the U-Net, the component with the highest VRAM usage. We observed that quantizing with INT4 precision led to a performance loss in the quantization library we utilized as seen in Fig.~\ref{fig:supp_int4_deg}. Consequently, we also noted a degradation in the performance of the personalized results when using this quantization approach. Nevertheless, our results demonstrate that ZOODiP is fully capable of personalizing 4-bit quantized diffusion models using only \textbf{1.9 GB} of memory during the entire training process. Fig.\ref{fig:supp_int4} and Tab.\ref{tab:supp_4bit} present the qualitative and quantitative results, respectively, for personalization with the 4-bit quantized model. To utilize the \texttt{qint4} data type, we employed BFloat16 (BF16) for activation in all units, including the VAE, U-Net, and text encoder. When using BF16, the default $\mu$ value of $10^{-3}$ was insufficient, so we adjusted it to $10^{-2}$ to achieve optimal performance.

\begin{table}[]
    \centering
    \begin{tabular}{c|c}
        \toprule
        Method & $\text{DINO}_{\text{inter}}\downarrow$ \\
        \midrule
        DB & 0.825 \\
        QLoRA & 0.731 \\
        PEQA & 0.806 \\
        TuneQDM & 0.778 \\
        TI & 0.679 \\
        GF-TI & 0.150 \\
        \textbf{ZOODiP (Ours)} & 0.671\\
        \bottomrule
    \end{tabular}
    \caption{Comparison of $\text{DINO}_{\text{inter}}$ scores across various personalization methods. The results indicate that TI-based methods are capable of generating a diverse range of images, whereas DB-based methods exhibit relatively lower diversity. This observation was consistent across all subjects in the DreamBooth (DB) dataset, highlighting a fundamental difference in the ability of these methods to capture and reflect variations in the generated outputs.}
    \label{tab:supp_diversity}
\end{table}

\begin{figure*}
  \centering
  \includegraphics[width=1.0\linewidth]{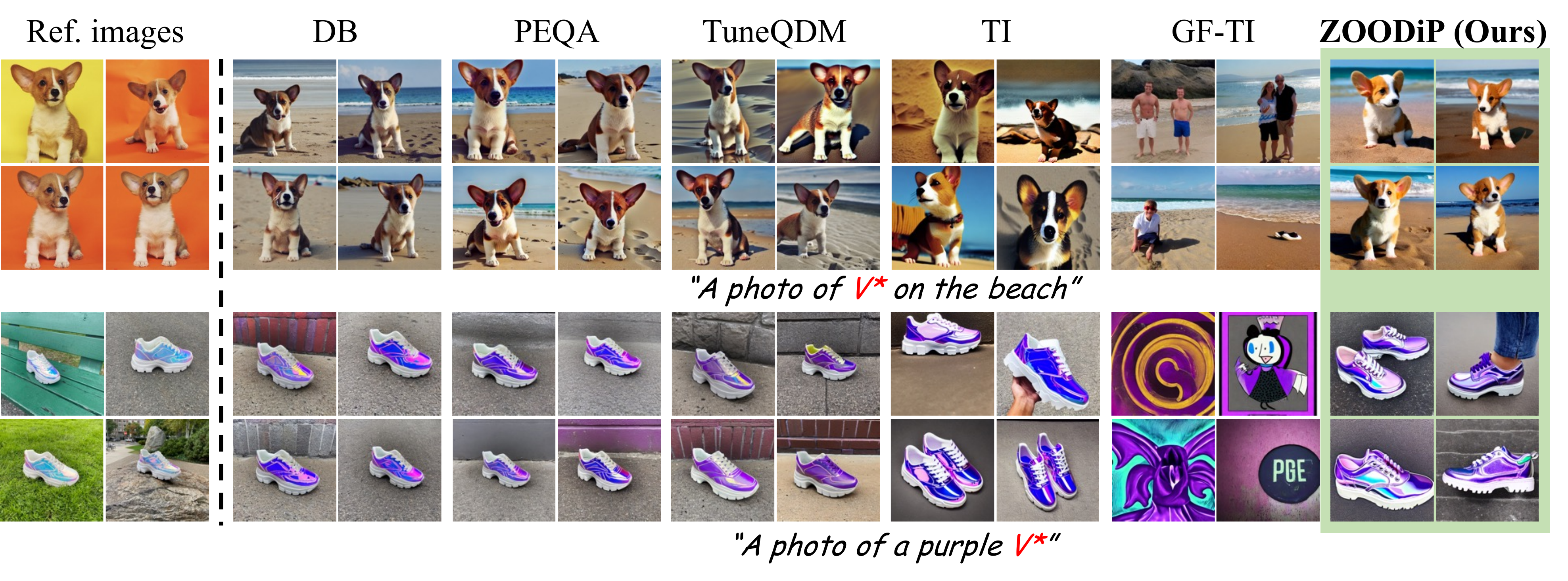}
  \caption{\textbf{Qualitative comparison of the diversity of generated images} This figure compares the diversity achieved by different personalization methods. ZOODiP demonstrates the ability to generate highly diverse images while utilizing minimal memory resources.}
  \label{fig:supp_diversity}
\end{figure*}

\begin{figure*}[h]
  \centering
  \includegraphics[width=1.0\linewidth]{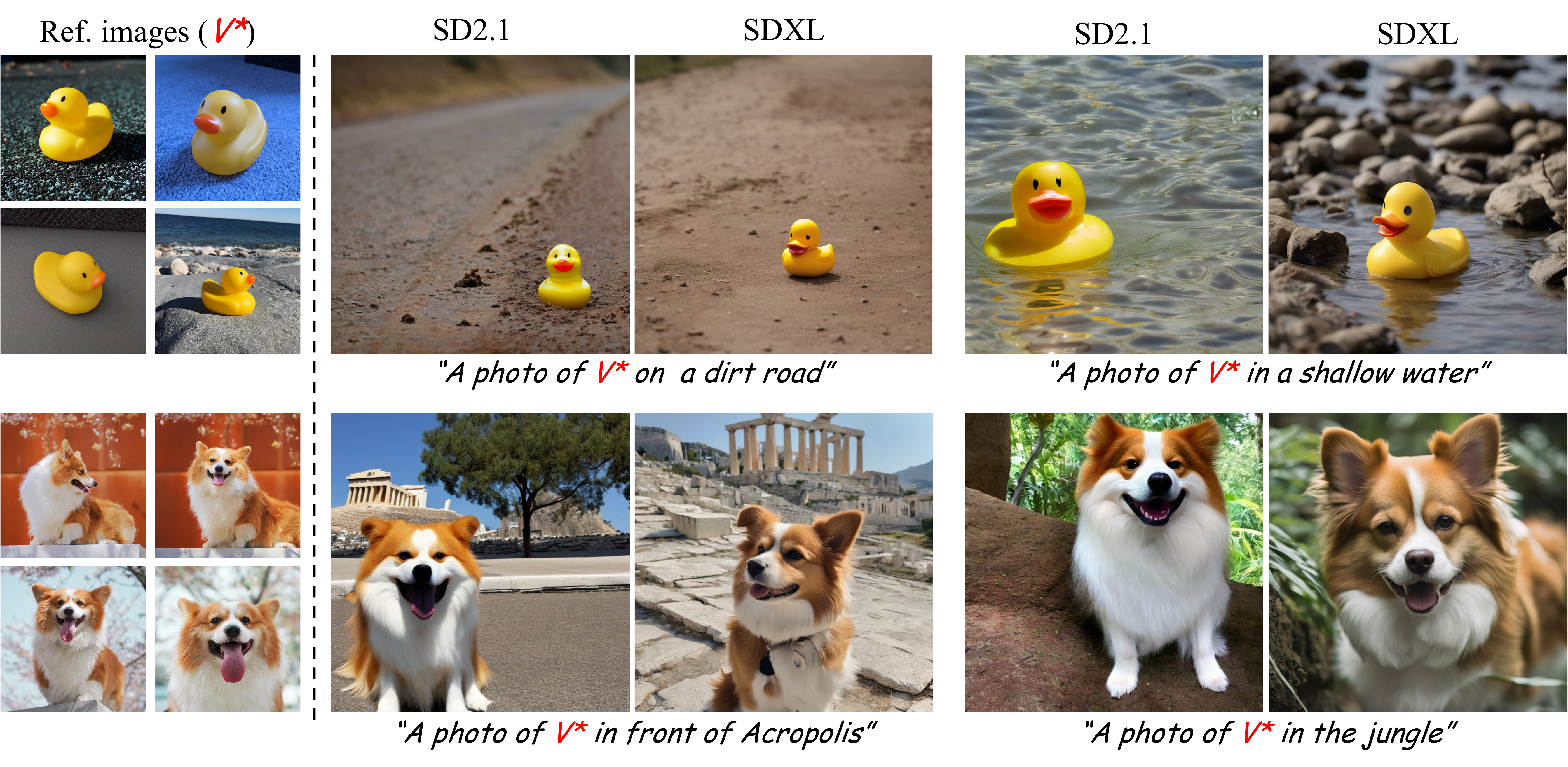}
  \caption{Qualitative results for personalizing SD2.1 and SDXL with ZOODiP. The figure demonstrate that ZOODiP can be applied not only to SD1.5, as discussed in the main paper, but also to various other models. For SD2.1, inference were conducted with images at a resolution of $768\times768$, while for SDXL, image generation was performed with resolution of $1024\times1024$. However, for SDXL, it was observed that the model's inherent color interpretation prevents the subject's colors from being completely replicated. This indicates that the model's color rendering can vary depending on the environmental context, leading to shifts in the perceived color scheme.}
  \label{fig:supp_generalize}
\end{figure*}

\subsection{Diversity of generated images}
TI-based methods tend to exhibit less overfitting compared to DreamBooth (DB)-based methods, resulting in higher diversity in the generated images.
To evaluate this tendency, we introduce the $\text{DINO}_{\text{inter}}$ score. For each subject, we generated 50 images using the same prompt with a personalized model. The DINOv2 embedding cosine similarity was then calculated for all pairs of generated images and averaged. A higher $\text{DINO}_{\text{inter}}$ score indicates that the generated images are more similar to each other, reflecting lower diversity in the generated outputs. The quantitative results for $\text{DINO}_{\text{inter}}$ are presented in the Tab.~\ref{tab:supp_diversity}. Additionally, we provide qualitative diversity results in Fig.~\ref{fig:supp_diversity}. Fig.~\ref{fig:supp_diversity} shows a comparison of diversity across different methods, clearly showing that TI-based methods achieve significantly higher diversity than other approaches.

\subsection{Generalizability to other models}
\label{supp_sec:general}
In the main paper, our experiments were conducted using Stable Diffusion v1.5 (SD1.5). However, ZOODiP is not limited to the certain model and can be extended to other models trained on different datasets and architectures, such as SD2.1 and SDXL~\cite{podell2023sdxl} as seen in Fig.~\ref{fig:supp_generalize}. For SD2.1, training was performed on a larger text token embedding dimension (1024 in \texttt{OpenCLIP-ViT-H}) using the same hyperparameters as SD1.5 except for $\mu=10^{-2}$ and $\eta=10^{-2}$. For SDXL, we maintained all hyperparameters except for $\mu=10^{-2}$, $T_L=700$, $L=20,000$ and trained both token embeddings of \texttt{OpenCLIP-ViT-L} (768 dimension) and \texttt{OpenCLIP-ViT-G} (1280 dimension). The adjustment of $T_L$ is supported by prior research which indicates that information loss vary with dimensionality~\cite{lin2024common, cross2023offset}, and the value of $T_L$ was selected empirically. Notably, full-precision (FP32) Textual Inversion on SDXL requires approximately 17 GB of memory, and DreamBooth cannot be trained on customer level GPUs, such as the RTX3090. In contrast, tuning SDXL with ZOODiP enables successful training of concepts with only \textbf{5 GB of memory}, highlighting its efficiency and scalability across diverse model configurations.

\begin{table}[]
    \centering
    \begin{tabular}{cc|c}
    \toprule
     Base. &  Method  & Time (min)  \\
    \midrule
    \multirow{4}{*}{DB} & DB & 2 \\
    & QLoRA & 1.1 \\
    & PEQA & 1.5 \\
    & TuneQDM & 1.4 \\
    \midrule
    \multirow{3}{*}{TI} & TI & 8.8 \\
    & GF-TI & 293\\
    & ZOODiP ($n=2$) & 31\\
    \bottomrule
    \end{tabular}
    \caption{Total training time with the configurations in Sec~\ref{sec:supp_config}. DB-based methods consume more memory but train faster, while TI-based methods are more memory-efficient but require more iterations.}
    \label{tab:supp_training_time}
\end{table}

\subsection{Training time}
We measured the total training time for each personalization method and observed a general inverse relationship between memory usage and training time in Tab.~\ref{tab:supp_training_time}. The table shows a trade-off between time complexity and space complexity. However, Gradient-Free Textual Inversion (GF-TI) deviates from this trend, requiring significantly more training time despite its low memory usage. This behavior indicates that GF-TI struggles with learning effectively under small batch training conditions, which impacts its overall efficiency and practicality in resource-constrained devices.

\begin{figure}[h]
  \centering
  \includegraphics[width=1.0\linewidth]{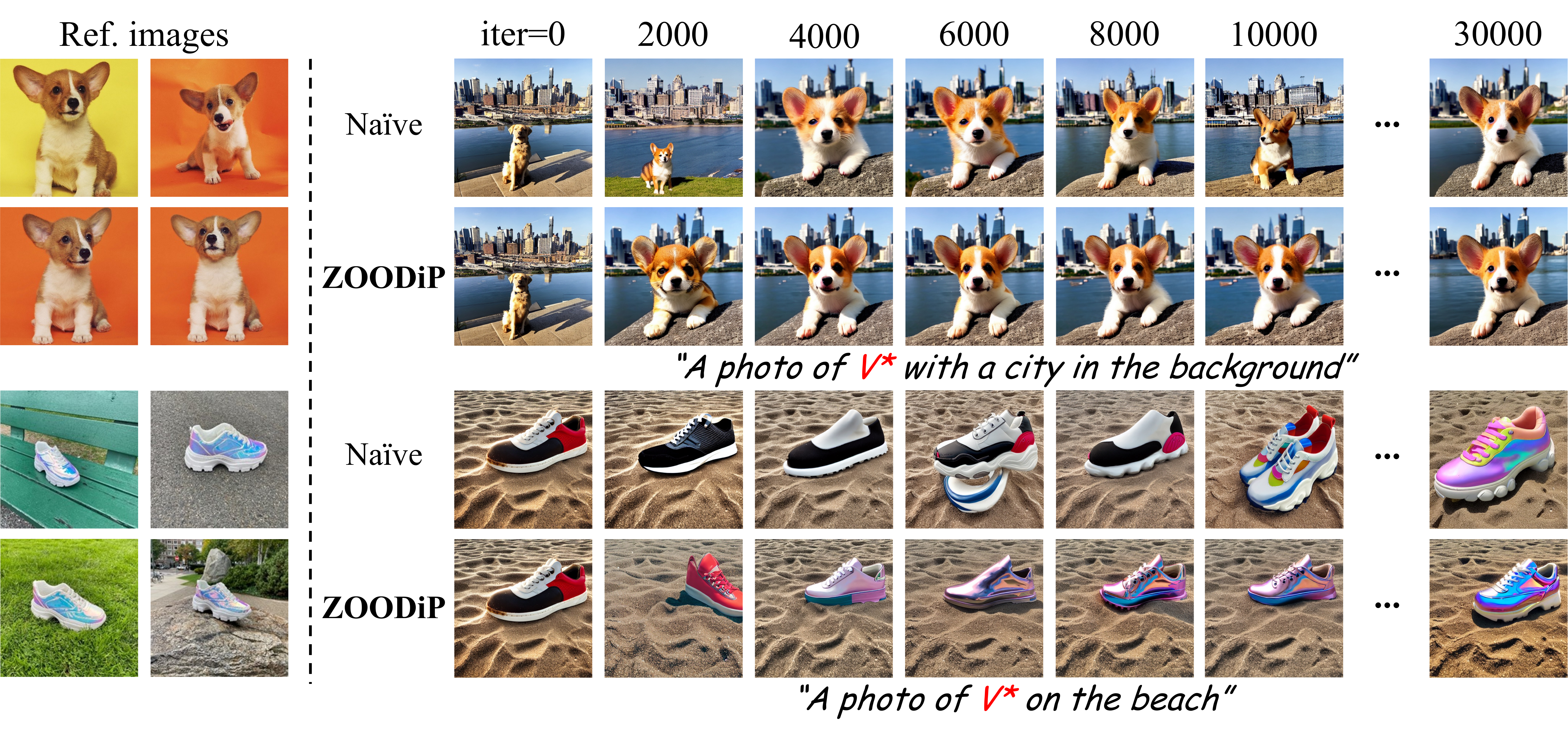}
  \caption{Qualitative comparisons on na\"ive ZO textual inversion without SG and PUTS to ZOODiP (Ours) over iterations. The na\"ive approach exhibits slower training and tends to produce images that are less aligned with the reference image. In contrast, ZOODiP achieves faster training and generates images that are closely aligned with the reference subject.}
  \label{fig:supp_sg_puts}
\end{figure}

\section{Additional Results}
\subsection{Qualitative results of SG and PUTS}
Subspace Gradient (SG) and Partial Uniform Timestep Sampling (PUTS) play crucial roles in enabling efficient personalization by addressing two key challenges: SG removes noisy gradient components to focus on the most informative subspace, while PUTS suppresses sampling from ineffective timesteps to optimize training efficiency. To evaluate the impact of these components during the learning process, we generated images using the same seed with token embeddings learned at every $1,000$ iterations.
As illustrated in Fig.~\ref{fig:supp_sg_puts}, using both SG and PUTS significantly accelerates the training process compared to na\"ive zeroth-order optimization that does not incorporate these enhancements. The results show that models using SG and PUTS converge to high-quality personalization much faster than the baseline, underscoring the effectiveness of these components in improving the efficiency and speed of training.

\subsection{Additional qualitative results}
\begin{figure*}[!p]
  \centering
  \includegraphics[width=1.0\linewidth]{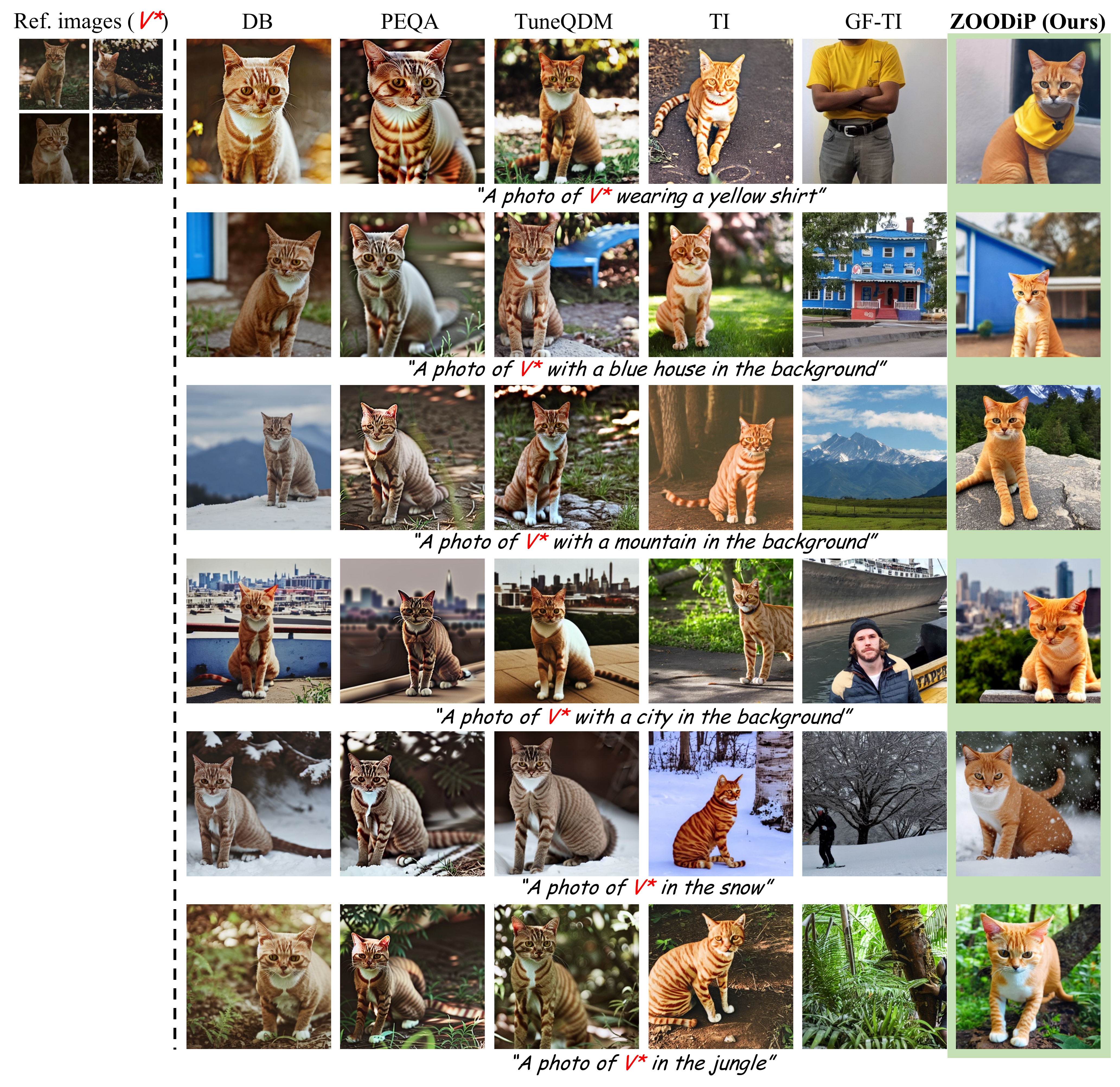}
  \caption{Qualitative comparison of image and text alignment on the \texttt{<cat>} subset of DB dataset.}
  \label{fig:supp_quali_cat}
\end{figure*}

\begin{figure*}[!p]
  \centering
  \includegraphics[width=1.0\linewidth]{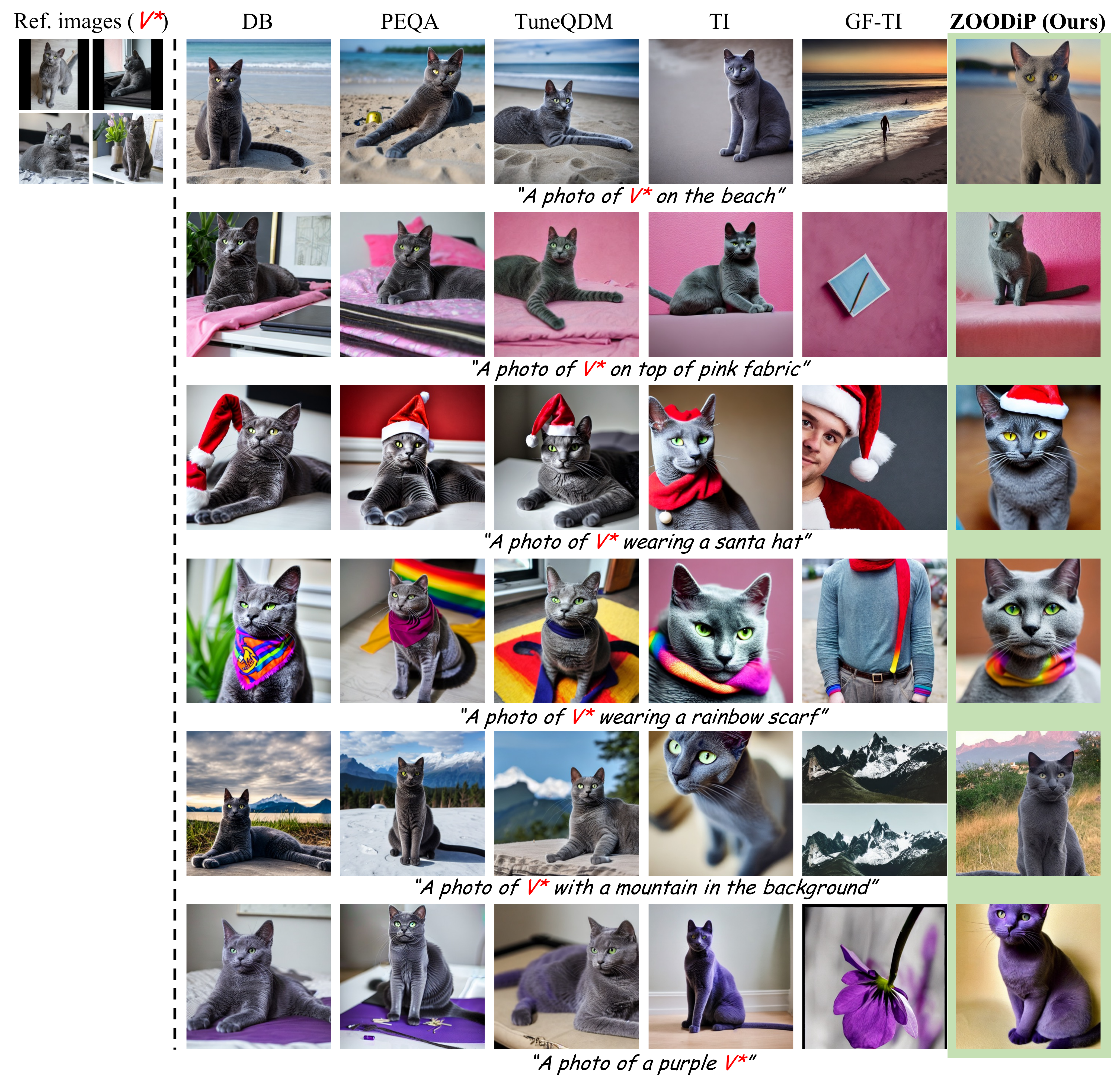}
      \caption{Qualitative comparison of image and text alignment on the \texttt{<cat2>} subset of DB dataset.}
  \label{fig:supp_quali_cat2}
\end{figure*}

\begin{figure*}[!p]
  \centering
  \includegraphics[width=1.0\linewidth]{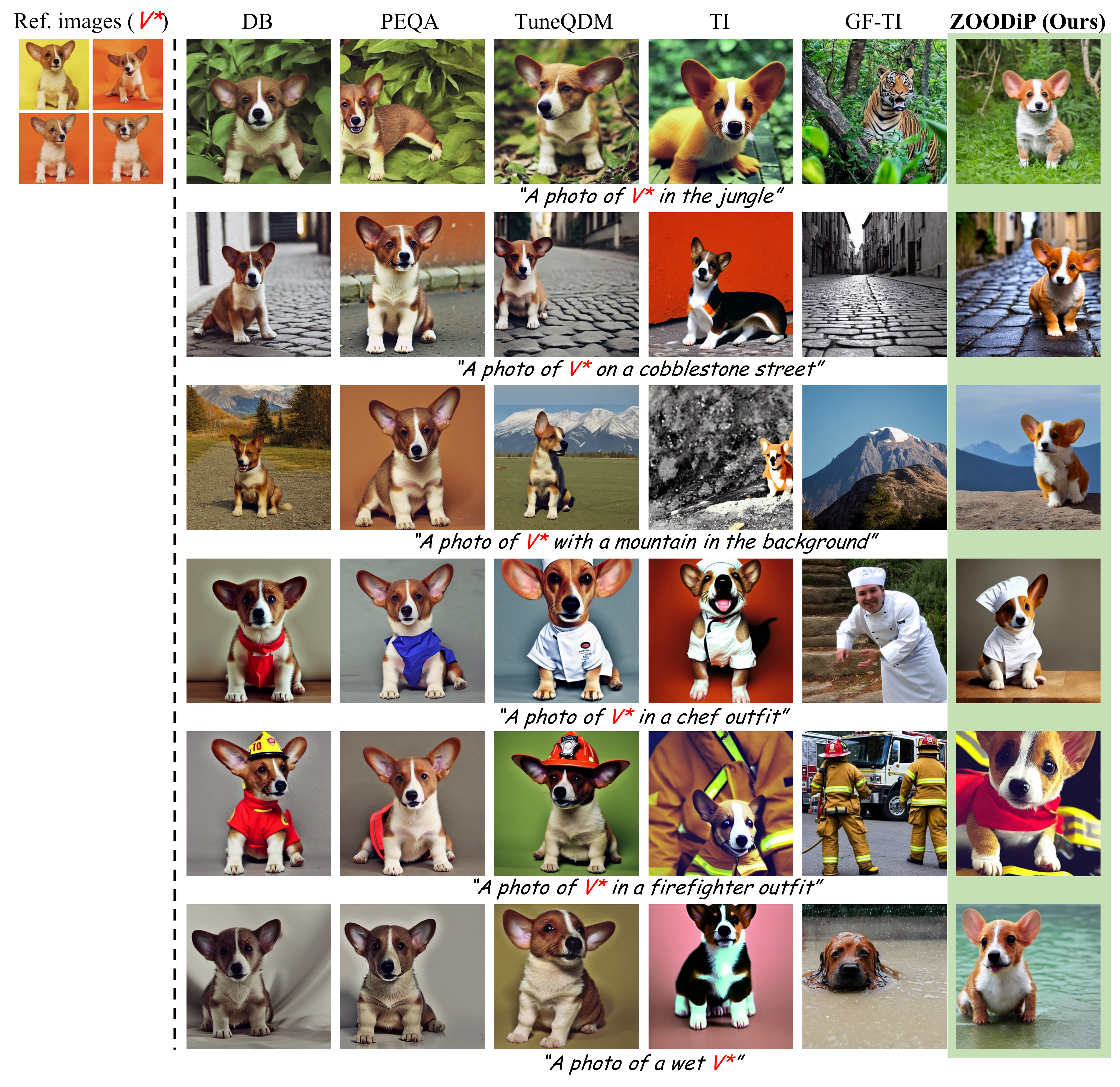}
  \caption{Qualitative comparison of image and text alignment on the \texttt{<dog6>} subset of DB dataset.}
  \label{fig:supp_quali_dog6}
\end{figure*}

\begin{figure*}[!p]
  \centering
  \includegraphics[width=1.0\linewidth]{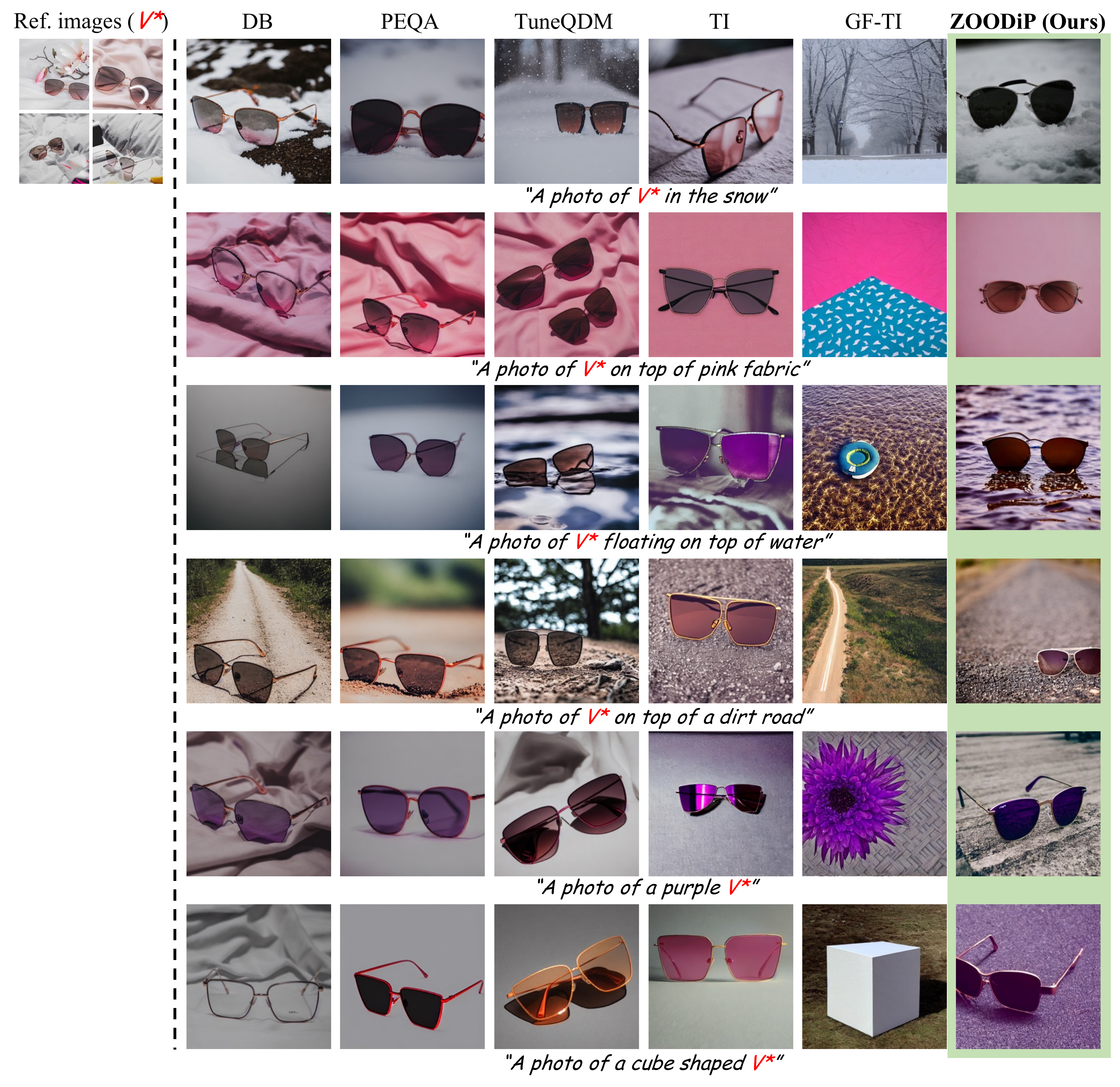}
  \caption{Qualitative comparison of image and text alignment on the \texttt{<pink\_sunglasses>} subset of DB dataset.}
  \label{fig:supp_quali_pink_sunglass}
\end{figure*}

\begin{figure*}[!p]
  \centering
  \includegraphics[width=1.0\linewidth]{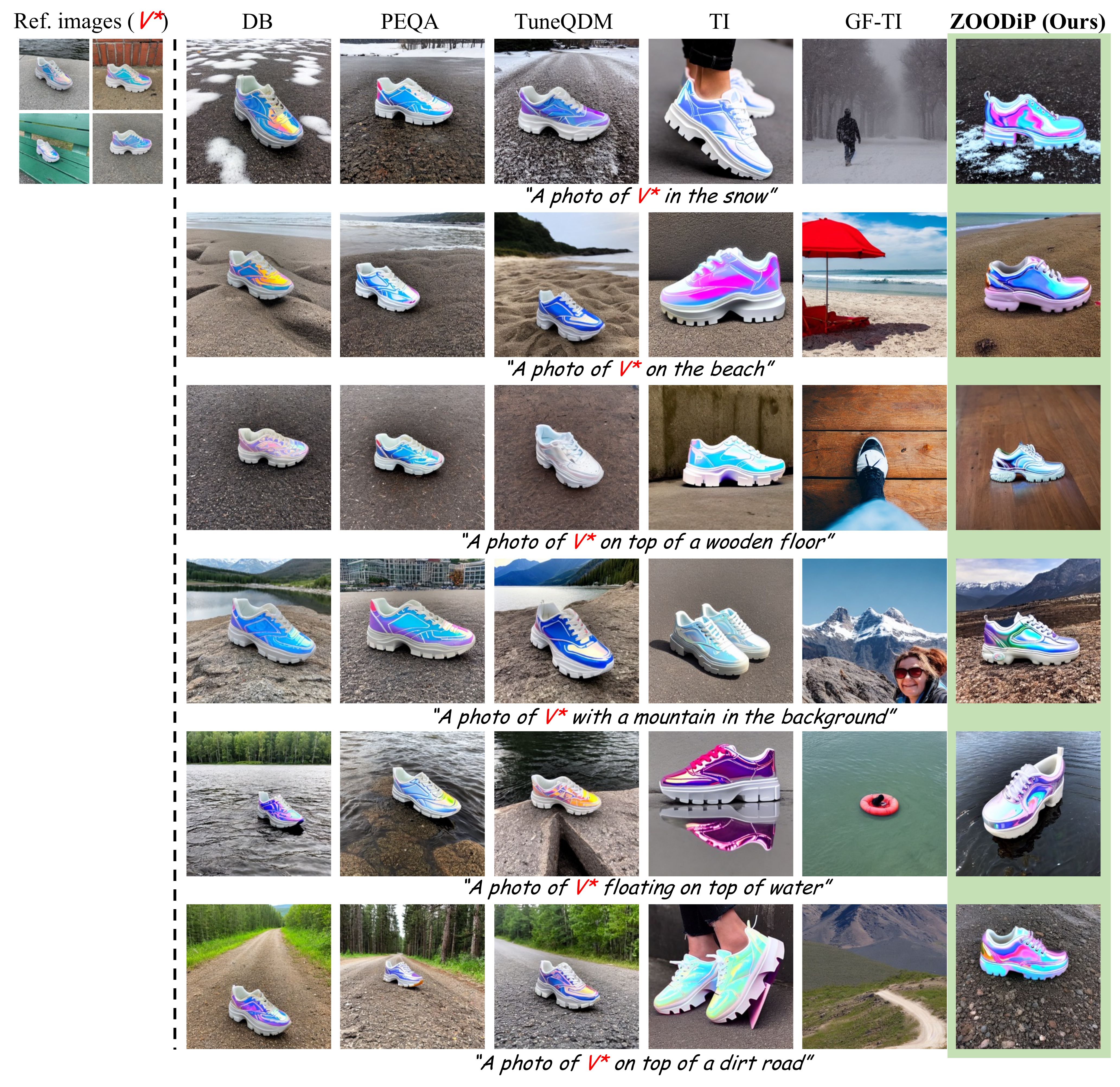}
  \caption{Qualitative comparison of image and text alignment on the \texttt{<shiny\_sneaker>} subset of DB dataset.}
  \label{fig:supp_quali_shiny_sneaker}
\end{figure*}

In this section, we present additional qualitative comparisons of personalization results that were not included in the main paper. These results are illustrated in Fig.~\ref{fig:supp_quali_cat}, Fig.~\ref{fig:supp_quali_cat2}, Fig.~\ref{fig:supp_quali_dog6}, Fig.~\ref{fig:supp_quali_pink_sunglass}, and Fig.~\ref{fig:supp_quali_shiny_sneaker}. The comparisons demonstrate that ZOODiP effectively captures the characteristics of the reference image and text prompts, achieving a level of fidelity comparable to other gradient-based personalization methods.

\section{Limitation}
In this section, we delve into the limitations of ZOODiP. While ZOODiP incorporates innovative techniques such as Subspace Gradient and Partial Uniform Timestep Sampling to mitigate the slow learning speed—a well-known challenge of zeroth-order optimization—it still requires a considerably larger number of iterations compared to gradient-based approaches. This increased iteration count stems from the fundamental nature of zeroth-order optimization, which relies on function evaluations rather than gradient backpropagation, inherently making it less sample-efficient. Although ZOODiP compensates with a faster training speed per iteration, the substantial number of iterations can introduce a significant time overhead, especially when proper hardware acceleration, such as high-performance Neural Processing Units (NPUs), is unavailable.

Furthermore, as ZOODiP is built upon the Textual Inversion framework, its performance is influenced by the strengths and weaknesses of Textual Inversion. This dependency implies that ZOODiP may face challenges in cases where Textual Inversion struggles, such as subjects with complex or highly variable visual characteristics, or when adapting to certain models that inherently perform poorly in personalization tasks. For example, if the base model lacks sufficient representational capacity or if the dataset used for training does not adequately capture the nuances of the subject, the effectiveness of ZOODiP can diminish.
\clearpage
  {
    \small
    \bibliographystyle{ieeenat_fullname}
    \bibliography{main}
}

\end{document}